\documentclass{article}

    \PassOptionsToPackage{numbers, compress}{natbib}

\usepackage[preprint]{neurips_2023}

\usepackage[utf8]{inputenc} 
\usepackage[T1]{fontenc}    
\usepackage{hyperref}       
\usepackage{url}            
\usepackage{booktabs}       
\usepackage{amsfonts}       
\usepackage{nicefrac}       
\usepackage{microtype}      
\usepackage{xcolor}         
\usepackage{blindtext}
\usepackage{graphicx}
\usepackage{subcaption}
\usepackage{amsmath}
\usepackage{multicol}
\usepackage{algorithm}
\usepackage{algpseudocode}
\usepackage{listings}
\usepackage{longtable}

\usepackage[export]{adjustbox}
\usepackage{graphics}
\usepackage{array}

\definecolor{codegreen}{rgb}{0,0.6,0}
\definecolor{codegray}{rgb}{0.5,0.5,0.5}
\definecolor{codepurple}{rgb}{0.58,0,0.82}
\definecolor{backcolour}{rgb}{0.95,0.95,0.92}

\lstdefinestyle{mystyle}{
    backgroundcolor=\color{backcolour},   
    commentstyle=\color{codegreen},
    keywordstyle=\color{magenta},
    numberstyle=\tiny\color{codegray},
    stringstyle=\color{codepurple},
    basicstyle=\ttfamily\footnotesize,
    breakatwhitespace=false,         
    breaklines=true,                 
    captionpos=b,                    
    keepspaces=true,                 
    numbers=left,                    
    numbersep=5pt,                  
    showspaces=false,                
    showstringspaces=false,
    showtabs=false,                  
    tabsize=2
}

\title{torchmSAT: A GPU-Accelerated Approximation To The Maximum Satisfiability Problem}

\author{%
  Abdelrahman Hosny \\
  Department of Computer Science\\
  Brown University\\
  Providence, RI 02902 \\
  \texttt{abdelrahman\_hosny@brown.edu} \\
  \And
  Sherief Reda \\
  School of Engineering \\
  Brown University \\
  Providence, RI, 02902 \\
  \texttt{sherief\_reda@brown.edu} \\
}

\begin{document}

\maketitle

\begin{abstract}
  The remarkable achievements of machine learning techniques in analyzing discrete structures have drawn significant attention towards their integration into combinatorial optimization algorithms. Typically, these methodologies improve existing solvers by injecting learned models within the solving loop to enhance the efficiency of the search process. In this work, we derive a single differentiable function capable of approximating solutions for the Maximum Satisfiability Problem (MaxSAT). Then, we present a novel neural network architecture to model our differentiable function, and progressively solve MaxSAT using backpropagation. This approach eliminates the need for labeled data or a neural network training phase, as the training process functions as the solving algorithm. Additionally, we leverage the computational power of GPUs to accelerate these computations. Experimental results on challenging MaxSAT instances show that our proposed methodology outperforms two existing MaxSAT solvers, and is on par with another in terms of solution cost, without necessitating any training or access to an underlying SAT solver. Given that numerous NP-hard problems can be reduced to MaxSAT, our novel technique paves the way for a new generation of solvers poised to benefit from neural network GPU acceleration.
\end{abstract}

\section{Introduction}
\label{sec:introduction}
In recent years, machine learning has emerged as a powerful method for solving complex combinatorial optimization problems, a class of problems that are notably difficult to solve due to their inherent combinatorial nature.
These problems often fall within the realm of Karp's NP-hard complexity class \citep{karp_reducibility_1972}, requiring a substantial amount of computational resources to find exact solutions as the problem size increases.
The application of machine learning, particularly deep neural networks, has demonstrated remarkable potential in tackling such problems \citep{bengio2021machine}.
Machine learning techniques encompass a diverse range of approaches, including those that substitute traditional heuristics within existing algorithms \citep{gasse2019exact, li2018combinatorial, wang2021bi, khalil2017learning}, enhance solvers by incorporating learned configurations \citep{hu2022branch, vaezipoor2021learning, selsam2019guiding}, or reformulate a problem as a Markov Decision Process and employ reinforcement learning strategies \citep{mazyavkina2021reinforcement, kool2018attention, khalil2017learning-rl, bello2016neural}. 
These techniques have been successfully employed across various domains, from operations research and graph theory to constraint satisfaction and scheduling problems, offering a new perspective on addressing the challenges posed by NP-hard problems \citep{bengio2021machine}.

Satisfiability (SAT) and Maximum Satisfiability (MaxSAT) problems represent two prominent classes of combinatorial optimization problems.
In the case of SAT, the objective is to determine whether there exists an assignment of truth values to a given set of Boolean variables that satisfies a given set of logical clauses, typically expressed in conjunctive normal form (CNF).
When the problem is non-satisfiable, MaxSAT extends the SAT problem by seeking an assignment that maximizes the number of satisfied clauses, rather than trying to satisfy all clauses.
MaxSAT problems are of particular interest due to their ability to model optimization scenarios where some degree of constraint violation is tolerable, capturing a broader range of real-world applications \citep{shabani2018pmtp, si2017maximum, ivanvcic2008efficient}.
Both SAT and MaxSAT problems have been extensively studied, giving rise to a variety of algorithmic techniques and heuristics aimed at efficiently solving or approximating solutions to these problems \citep{ansotegui2013sat}.

Existing MaxSAT solvers employ a diverse range of techniques to efficiently tackle the Maximum Satisfiability problem.
These techniques can generally be classified into two main categories: \textit{complete} and \textit{incomplete} methods.
Complete methods, such as Branch and Bound algorithms \citep{li2022boosting}, search exhaustively for an optimal solution, guaranteeing the best possible assignment of truth values to satisfy the maximum number of clauses.
Some complete solvers utilize SAT solvers as a core engine, iteratively refining their search space by tightening the upper bound of unsatisfied clauses, as seen in iterative SAT-based MaxSAT algorithms \citep{morgado2014core, morgado2014mscg}.
In contrast, incomplete methods, like local search algorithms, do not guarantee optimality but aim to find high-quality solutions in less time.
These methods involve exploring the solution space by iteratively making small changes to the current assignment of truth values, guided by various heuristics and neighborhood search strategies \citep{martins2014incremental, morgado2013iterative}.
Both complete and incomplete MaxSAT solvers have been successfully applied to real-world problems, with the choice of solver being dependent on the problem size, required optimality guarantees, and available computational resources.

\begin{figure}
    \centering
    \begin{subfigure}{0.5\textwidth}
        \includegraphics[clip, trim=3cm 0 4cm 0, scale=0.6]{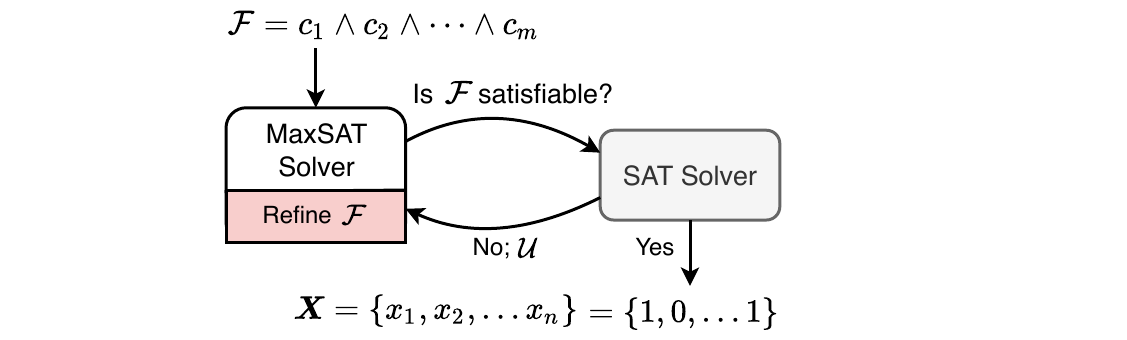}
        \caption{Existing MaxSAT Solvers}%
    \end{subfigure}%
    \begin{subfigure}{0.5\textwidth}
        \includegraphics[clip, trim=2cm 0 1cm 0, scale=0.6]{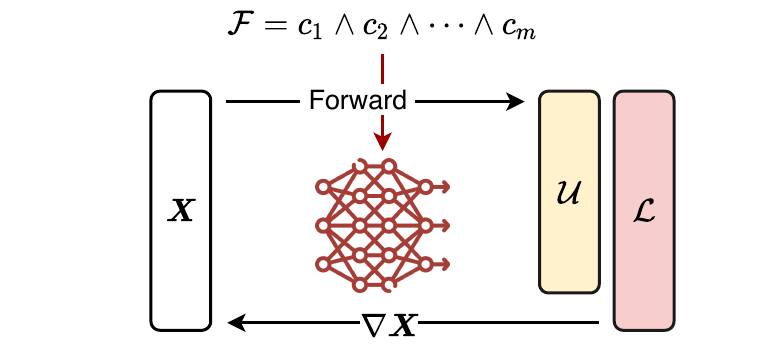}
        \caption{Our torchmSAT Solver}
    \end{subfigure}
  \vspace{-0.2in}
  \caption{Existing MaxSAT solvers depend on a SAT oracle to iteratively evaluate a Boolean formula, $\mathcal{F}$, and update the clauses in $\mathcal{F}$ to reduce the number of unsatisfied clauses, $\mathcal{U}$. Our torchmSAT solver eliminates the need for a SAT oracle and encodes $\mathcal{F}$ in the architecture of a neural network. Maximizing satisfiability is performed using backpropagation on the neurons contributing to $\mathcal{U}$.}
  \label{fig:motivation}
  \vspace{-0.2in}
\end{figure}

In this paper, we present a novel MaxSAT solver, called \textit{torchmSAT}, that leverages neural networks and falls within the incomplete category.
Departing from the conventional practice of incrementally enhancing existing SAT solvers for MaxSAT resolution, we propose a completely new algorithm developed from scratch.
As depicted in Figure \ref{fig:motivation}, our method forgoes the necessity for a traditional SAT solver as a fundamental component of the search algorithm in incomplete techniques for MaxSAT.
Rather than training a neural network to predict assignments that maximize satisfied clauses, an inherently complex task, we develop a novel neural network architecture with a differentiable loss function for solving MaxSAT.
The key intuition is that by relaxing the binary constraint of the problem and allowing the Boolean variables to be represented in a continuous domain, advances in deep learning libraries would allow this optimization to be executed efficiently.
Accordingly, the training process can be seen as a process that iteratively explores the solution space, generating progressively improved variable assignments.
Consequently, our approach eliminates the need for labeled training data, or the need to call an underlying SAT solver to testify (un)satisfiability.
Moreover, since our proposed solver is natively built using a reliable deep learning library \citep{paszke2019pytorch}, we are able to run the solver on GPUs without any change to the data structure, the solving algorithm or the optimization process.
Therefore, we investigate the advantages of hardware acceleration for solving MaxSAT, demonstrating that the acceleration of such computations enables the solver to traverse the feasible solution space more rapidly.
In essence, torchmSAT presents a fresh approach for solving MaxSAT that could potentially open doors for a new generation of combinatorial optimization solvers.

\section{Preliminaries}
\label{sec:preliminaries}

\textbf{MaxSAT Formulation.}
In propositional logic, a Boolean formula is composed of Boolean variables, $\mathbf{x} = \{x_1, x_2, \dots x_n\}$, and logical operators, including negations ($\neg$), conjunctions ($\wedge$), and disjunctions ($\vee$).
A common representation for Boolean formulas is the conjunctive normal form (CNF), which is structured as a conjunction of multiple clauses, $\boldsymbol{C} = \{c_1, c_2, \dots, c_m \}$.
Each clause is a disjunction of literals, where a literal can be either a variable or its negation.

\vspace{-0.25in}
\begin{align*}
\mathcal{F} &= c_1 \wedge c_2 \wedge \dots \wedge c_m \\
\end{align*}

\vspace{-0.55in}

\begin{align}
\mathcal{F} &= (x_{1,1} \vee x_{1,2} \vee \dots \vee x_{1, n_1}) \wedge (x_{2,1} \vee \dots \vee x_{2,n_2}) \wedge  \dots \wedge{} (x_{m,1} \vee \dots \vee x_{m,n_m})
\label{eq:cnf}
\end{align}

This format facilitates the systematic analysis and manipulation of Boolean expressions for various computational tasks.
A formula is deemed satisfiable if there is at least one assignment of Boolean variables, $\mathbf{}{x}$, that satisfies all clauses. In numerous applications, a formula may not be entirely satisfiable, and the objective of MaxSAT solvers is to identify an assignment that satisfies the greatest number of clauses. In this scenario, the number of unsatisfied clauses is referred to as \textbf{the cost} of the CNF \--- a lower cost corresponds to a better assignment.
In some applications, (weighted) partial CNF formulas are considered.
Clauses in a partial CNF formula are characterized as hard, $\mathcal{H}$, meaning that these must be satisfied, or soft, $\mathcal{S}$, meaning that these are to be satisfied, if at all possible.

\textbf{SAT Oracle.} 
Existing MaxSAT solvers employ SAT oracles to handle CNF formulas.
A SAT oracle is any algorithm that can determine the satisfiability of any given Boolean formula in the conjunctive normal form (CNF).
If the formula is satisfiable, the SAT oracle provides a satisfying assignment; if not, it returns additional information such as an unsatisfiable core, $\mathcal{U} \subseteq \mathcal{F}$.
The unsatisfiable core, $\mathcal{U}$, represents a subset of $\mathcal{F}$'s clauses that are inherently unsatisfiable.
The task of determining the satisfiability of a given CNF is indeed NP-complete \citep{karp_reducibility_1972}.

\textbf{Conflict-Driven Clause Learning (CDCL)}
is a method used by modern SAT solvers to efficiently search the solution space and resolve conflicts during the search process.
State-of-the-art SAT solvers build upon the basic DPLL (Davis-Putnam-Logemann-Loveland) algorithm \citep{nieuwenhuis2006solving}, which employs backtracking and unit propagation, by adding clause learning and non-chronological backtracking \citep{biere2009handbook}.
When a CDCL-based solver encounters a conflict, which occurs when the current partial assignment of $\boldsymbol{X}$ leads to an unsatisfiable core $\mathcal{U}$, it analyzes the conflict to generate a new learned clause.
This learned clause represents the assignments of variables contributing to the conflict and helps in preventing the solver from introducing similar conflicts in the future.
The solver then performs jumping back several levels in the search tree instead of just one, using the learned clause to guide the process.
The main advantage of CDCL-based solvers is their ability to learn from conflicts and use that knowledge to prune the search space more effectively.
This results in faster and more efficient SAT solving for many real-world problem instances.

\textbf{MaxSAT Solvers.}
In an unweighted MaxSAT problem, a typical MaxSAT algorithm proceeds by making several calls to an underlying SAT oracle such as CaDiCaL \citep{fleury2020cadical}, Glucose \citep{audemard2018glucose} or Minisat \citep{sorensson2005minisat}.
The difference amongst algorithms is how they orchestrate the calls to the SAT solver.
For example, the RC2 algorithm sends $\mathcal{F}$ to the SAT solver, which reports that $\mathcal{F}$ is unsatisfiable and returns an unsatisfiable core, $\mathcal{U}$.
At least one of the clauses of the core will have to be disregarded in order to fix the core.
So, the algorithm proceeds by relaxing each clause in $\mathcal{U}$ (i.e. augmenting the clause with a fresh variable called relaxation variable) and constrains the sum of the relaxation variables to be at most one \citep{ignatiev2019rc2}.
In other words, the algorithm starts by assuming all clauses can be satisfied and iteratively relaxes this assumption until it finds a satisfying assignment.
Similarity, the FM algorithm \citep{manquinho2009algorithms} involves a sequence of calls to an unsatisfiability oracle, each of which generates an unsatisfiable core.
The clauses that are part of this unsatisfiable core are then relaxed, followed by the introduction of a new constraint that pertains to the relaxation variables in the formula.
On the other hand, the LSU algorithm runs a series of satisfiability oracle calls refining an upper bound on the MaxSAT cost, followed by one unsatisfiability call, which stops the algorithm \citep{morgado2013iterative}.
In other words, it finds an initial assignment with a suboptimal cost, and iteratively search for better assignments to reduce the cost.

In our work, we present a novel MaxSAT algorithm that does not rely on a SAT oracle during the search process. Instead, it adopts a progressive strategy akin to the one described in \citep{morgado2013iterative} and relies entirely on the neural network for identifying unsatisfiable cores, $\mathcal{U}$, and discovering improved solutions through backpropagation.

\section{Related Work}
\label{sec:related-work}
\textbf{ML for MaxSAT.}
Several recent efforts have investigated the integration of machine learning in solving SAT and MaxSAT problems.
\citet{selsam2018learning} train a graph neural network classifier to predict satisfiability of random SAT problems. 
The model learns to search for satisfying assignments during inference for problem instances that are larger than the ones seen during training.
Similarily, \citet{li_nsnet_2022} present a framework for SAT solving utilizing Belief Propagation (BP).
They introduce a Graph Neural Network (GNN) architecture that embeds BP in the latent space and use the trained model for marginal inference to obtain satisfying assignments for SAT.
Other learning-based methods have been proposed for MaxSAT.
For example, \citet{kumar2023learning} propose a method for learning combinatorial optimization problems from contextual examples, which indicate whether solutions are adequate in specific contexts.
The framework uses the MaxSAT formulation and considers a specific setting where example solutions and negative solutions are context-specific.
In the same way, \citet{berden2022learning} train MaxSAT models from examples and use a genetic algorithm that decreases the number of evaluations needed to find good models.
\citet{marino2021learning} uses a supervised learning approach to develop an algorithm that can fix Boolean variables based on local information from the Survey Propagation algorithm.
In general, this line of research collects training data on MaxSAT instances, and trains a model that generalizes for bigger problems.
Our method is different as it does not necessitate the gathering of any training data.

\textbf{NN-based Combinatorial Optimization.}
An emerging approach to incorporating machine learning for tackling combinatorial optimization problems involves regarding the neural network's training process as the problem-solving procedure for a given instance.
In this paradigm, a neural network is dynamically generated based on the problem instance, and through multiple iterations of forward and backpropagation, a series of learnable parameters embody the ultimate solution to the problem instance.
The absence of data requirements for training makes this method particularly attractive for deployment in diverse settings.
For instance, \citet{alkhouri2022differentiable} introduce a technique that operates on graphs and addresses the Maximum Independent Set (MIS) problem \citep{tarjan1977finding}. The core concept involves representing the MIS problem as a single differentiable function, which facilitates the utilization of differentiable solutions.
\citet{schuetz2022combinatorial} present a Graph Neural Network (GNN)-based solver for approximately solving combinatorial optimization problems by encoding the optimization problem using a Hamiltonian (cost function) and associating binary decision variables with vertices in an undirected graph.
A relaxation strategy is applied to generate a differentiable loss function for learning the GNN's node representations.
After several iterations, a softmax activation is applied to obtain one-dimensional probabilistic node assignments, which are then mapped back to integer variables using a projection heuristic.
Both work can be viewed as a Linear Programming (LP) relaxation of the Mixed-Integer Linear Programming (MILP) formulation of the investigated problems, i.e. Maximum Independent Set (MIS), and Maximim Cut (MaxCUT).
While our approach is inspired by the same ideas, we address a different problem (i.e. MaxSAT), and eliminate the use of GNNs.

\section{Methodology}
\label{sec:methodology}
\textbf{Key Idea.}
Our proposed method, \textit{torchmSAT}, is conceptually inspired by the technique of Linear Programming (LP) relaxations often utilized in the field of Mixed Integer Linear Programming (MILP) \citep{agmon1954relaxation}.
In MILP, certain variables are constrained to take only integer values which makes the optimization problem NP-hard.
A common strategy to tackle this issue is to apply LP relaxation, where the integer constraints on the variables are relaxed, allowing them to take on continuous values.
Similarly, in our method, we treat the binary variables of the MaxSAT problem as continuous, allowing us to leverage the power of differentiable optimization methods.
The challenge we try to address becomes constructing \textit{a single differentiable function} capable of approximating solutions for MaxSAT problems.
Once derived, implementing the optimization process using a neural network allows us to capitalize on existing deep learning libraries, and their acceleration capabilities.

In what comes next, we begin by deriving a single differentiable function for MaxSAT using a novel neural network architecture.
We describe the key components, their interactions, and their roles in the solving process.
Following this, we show how our solver is different in finding unsatisfiable cores and how the solving process proceeds.

\begin{figure}
  \centering
  \includegraphics[clip, trim=1.2cm 0 0 0, scale=0.9]{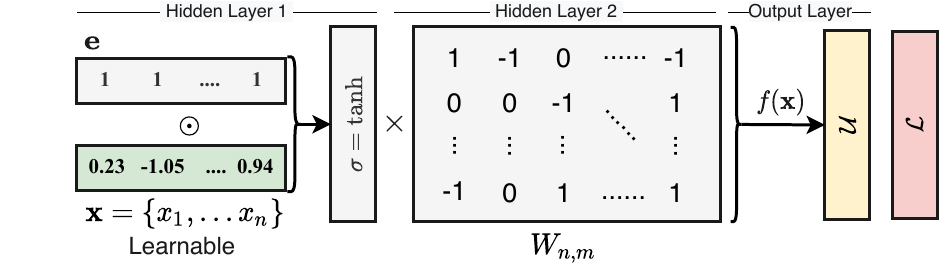}
\caption{Overview of torchmSAT neural network architecture. The learnable vector $\mathbf{x}$ represents the assignments of the Boolean variables in a conjunctive normal form, $\mathcal{F}$. The $W$ matrix is fixed and encodes a given MaxSAT instance represented in conjunctive normal form (CNF), i.e. Equation \ref{eq:cnf}, where the rows represent boolean variables, and columns represent clauses. A value\nobreakspace of\nobreakspace 1\nobreakspace or\nobreakspace-1 is assigned if a variable $x_i$ or $\neg x_i$ appears in clause $c_j$ respectively; and 0 otherwise. The output layer calculates the unsatisfied cores, $\mathcal{U}$. The loss function, $\mathcal{L}$, calculates gradients with respect to elements of $\mathbf{x}$ that are contributing to $\mathcal{U}$. The neural network requires no data for training. Rather, the training process functions as the solving process of maximizing the number of satisfied clauses in $\mathcal{F}$.}
\vspace{-0.1in}
\label{fig:torchmsat-architecture}
\end{figure}

\textbf{The Neural Network Architecture.}
Our proposed single differentiable function for MaxSAT is modeled as a neural network.
Figure \ref{fig:torchmsat-architecture} presents a high-level overview of its architecture.
At the heart of our solving algorithm lies a vector, $\mathbf{x}$, which represents the relaxations of the Boolean variables in a given MaxSAT instance (refer to Equation \ref{eq:cnf}).
The vector $\mathbf{x}$ is the only trainable set of parameters within the neural network.
Although $\mathbf{x}$ consists of real numbers, at any given point in time during the solving process, we can project it back into the binary domain to derive the variable assignments.
To reverse the relaxation, we interpret $x_i = 1$ when $x_i > 0$, and $0$ otherwise.
At the onset of the solving process, $\mathbf{x}$ is initialized with random real values.
Interpreting the values at this point merely corresponds to assigning random Boolean values to the variables.
Throughout the solving process, the values of $\mathbf{x}$ are incrementally updated in a direction that minimizes the loss.
This specific design prompts the network to progressively learn to satisfy an increasing number of clauses.

In the context of backpropagation and the chain rule, the derivative of a multiplication operation with respect to its inputs distributes the gradients.
Therefore, the first hidden layer performs a point-wise multiplication between $\mathbf{x}$ and a vector of identical length containing all values set to 1, denoted as\nobreakspace $\mathbf{e}$.
This vector serves to propagate the gradient to the corresponding $x_i$s during the backpropagation process.
After that, and to stabilize backpropagation, a $\tanh$ activation function is applied to constrain the values of the first layer within the range of -1 to 1.
The $\tanh$ activation prevents the output of the first layer from reaching excessively high or low values, which could potentially terminate the learning (i.e. solving) prematurely.
The rest of the neural network is fixed (i.e. non-learnable) and architected to encode our novel single differentiable function.

The second layer uniquely encodes a given MaxSAT instance.
A matrix $W_{n,m}$ is initialized at the start of the solving process, with rows representing variables $x_i$'s and columns representing clauses\nobreakspace $c_j$'s.
A value of $1$ is set if a variable $x_i$ appears in clause $c_j$, while a value of $-1$ is set if its negation $\neg x_i$ appears in $c_j$.
All other entries in the matrix are set to 0.
This formulation results in our unique differentiable function, $f(\mathbf{x})$, which generates a vector $\mathcal{U}$ of length $m$:

\vspace{-0.25in}
\begin{align*}
\mathcal{U} = f(\mathbf{x}) = \tanh(\mathbf{e} \odot \mathbf{x}) \cdot W
\end{align*}

where $\odot$ is element-wise multiplication, and $\cdot$ is a matrix multiplication. The entries of this vector, $\mathcal{U}$, indicate the (un)satisfiability status of each clause, serving as a dual-purpose: an evaluator for SAT and a guide for variable gradients.

\textbf{Example.}
Consider the following MaxSAT problem in CNF:

\vspace{-0.25in}
\begin{align*}
\mathcal{F} &= (\neg x_1) \wedge (\neg x_2) \wedge (x_1 \vee x_2)
\label{eq:cnf}
\end{align*}

This formula is not satisfiable.
It is evident that any combination of binary assignments for $x_1$ and $x_2$ can satisfy, at most, two clauses.
Assume that $\mathbf{x}$ is initialized randomly as $[0.43, 1.27]$, which is interpreted as $x_1=1$ and $x_2=1$ since both values are positive.
The output of the forward pass is:

\vspace{-0.2in}
\begin{equation*}
f(\mathbf{x}) = \tanh{(\left[\begin{array}{cc}
1 & 1
\end{array}\right] \odot \left[\begin{array}{cc}
0.43 & 1.27
\end{array}\right])} \cdot \left[\begin{array}{ccc}
-1 & 0 & 1 \\
0 & -1 & 1
\end{array}\right] = \left[\begin{array}{ccc}
-0.41 & -0.85 & 1.26
\end{array}\right]
\
\end{equation*}

Similar to the reverse relaxation operation of $\mathbf{x}$, the output of $f(\mathbf{x})$ is interpreted similarly, i.e. $c_j$ is satisfied if its activation in $f$ is positive, and vice versa.
In this case, the output indicates that the first two clauses are unsatisfied, while only the last clause is satisfied.
This represents the first role of $f(\mathbf{x})$ as a SAT evaluator.
To satisfy the first two clauses, the values of $\mathbf{x}$ need to become negative so that the activations of the first two clauses in $f$ are positive.
This is where the role of the loss function comes in.
The loss function, $\mathcal{L}$, computes the MSE between $f(x)$ and a zero vector for the unsatisfied clauses.
A single iteration of backpropagation updates the values of $x_1$ and $x_2$ in the negative direction of the gradient, moving their values closer to zero.
For instance, with a learning rate of $0.1$, the new values of $\mathbf{x}$ will be $[0.34, 1.23]$.
The process continues as the forward-loss-backward loop attempts to satisfy the first two clauses.
However, once the first two clauses are satisfied (i.e., $\mathbf{x}$'s are negative, representing a Boolean zero when reversing the relaxation), the third clause becomes unsatisfied.
This is when the training process explores other ways of satisfying more clauses.

\textbf{Unsatisfiable Cores}
Contrary to conventional MaxSAT solvers discussed in Section \ref{sec:related-work} that rely on a SAT oracle to identify the set of unsatisfied clauses, $\mathcal{U}$, if the formula is determined to be unsatisfiable, our proposed neural network model directly infers the unsatisfied clauses from $f(\mathbf{x})$.
Nevertheless, for larger problems, the activations of $f(\mathbf{x})$ alone might not be enough to ascertain satisfiability.
The reason is that the matrix multiplication occurring in the first two layers is a non-injective operation.
This implies that distinct assignments for $\mathbf{x}$ might yield identical activations for $f(x)$.
For instance, the inputs $\mathbf{x} = [0.43, -1.27]$ and $\mathbf{x} = [-1.27, 0.43]$ would result in the same negative activation for the aforementioned third clause, despite the fact that the de-relaxed variable assignments are totally different; one corresponds to $[1, 0]$ while the other corresponds to $[0, 1]$.
To overcome this, we use an alternative method to establish (un)satisfiability of clauses.
We construct a vector, $\mathbf{s}$, of length $m$, where each entry corresponds to the negative of the number of variables present in its respective clause, i.e., $s_j = - len(c_j)$.
We also project $\mathbf{x}$ onto its Boolean domain, such that $x_i' = 1$ if $x_i > 1$, and $0$ otherwise.
This makes a clause in $\mathcal{U}$ unsatisfiable \textit{if and only if} $\mathbf{x}' \cdot W = \mathbf{s}$.
We use this projection to mask out satisfied clauses in the loss function, and only calculate it for unsatisfied clauses.
Consequently, gradients are computed exclusively for those variables that contribute to the unsatisfied clauses, resulting in a more efficient optimization.

\textbf{The Solving Process.}
In contrast to conventional MaxSAT solvers, which necessitate invoking a SAT oracle, pausing to await a result, and then coordinating the subsequent call, our algorithm operates differently.
torchmSAT operates in a progressive manner, which implies that it incrementally improves the solutions as the search process unfolds.
Algorithm \ref{algo1} shows the main steps of our solving process.
The solving process is an iterative procedure that alternates between a forward pass, loss calculation, and backward propagation.
In Line 1, we construct the neural network layers based on the given problem instance.
While the time limit has not been exhausted, the forward pass, in Line 3, takes the current assignments of the Boolean variables, $\mathbf{x}$, and computes the output of the network, $f(\mathbf{x})$.
In addition, it calculates the reversed relaxation of $\mathbf{x}$, denoted as $\mathbf{x}'$, which is used to determine clauses are currently satisfied by the assignment in Line 4.
In Lines 5-7, if the current assignment satisfies more clauses than previously found solutions, it outputs this result and updates the value of the best cost (the lower, the better).
In Line 8, the loss function calculates the MSE between $f(\mathbf{x})$ and a zero vector of the same length, reflecting the number of unsatisfied clauses.
The backpropagation step, in Line 9, adjusts the values of $\mathbf{x}$ according to the gradient calculated by backpropagation, moving the assignments in a direction that reduces the number of unsatisfied clauses.
This process continues until all clauses are satisfied, as indicated by a zero loss, or a given time limit is reached.

\section{Experiments}
\label{sec:expr}

In this section, we present a series of experiments designed to evaluate the performance of our proposed algorithm and compare it with existing state-of-the-art MaxSAT solvers.
Our experiments aim to demonstrate the effectiveness of the method in various scenarios and showcase its strengths and limitations.
We describe the experimental setup, including the problem instances used, the choice of benchmarks, and the evaluation metrics employed.
Additionally, we provide a detailed analysis of the experimental results, highlighting key observations and insights.
Through these experiments, we aim to validate the applicability of our approach and its potential impact on combinatorial optimization.

\begin{algorithm}[t!]
\caption{torchmSAT Algorithm for Solving MaxSAT}\label{algo1}
\textbf{Input:} MaxSAT instance in conjunctive normal form (CNF). See Equation \ref{eq:cnf}\\
\textbf{Parameters:} Time limit ($T$) \\
\textbf{Output:} Assignment for Boolean variables ($\mathbf{x}$) that maximizes satisfiability
\begin{algorithmic}[1]
\State Init $\mathbf{x}$ randomly. Construct $W$ from CNF. Init $\mathbf{s}$ where $s_j = - len(c_j)$. Set best\_cost = \#clauses.
\While {current solving time < $T$}
\State Run forward pass, $f(\mathbf{x})$. Calculate $\mathbf{x}'$ where $x_i' = 1$ if $x_i > 1$, and $0$ otherwise. 
\State Calculate unsatisfied clauses, $\mathcal{U} = \mathbf{x}' \cdot W == \mathbf{s}$.
\If {\# unsatisfied clauses ($\mathcal{U}$) < best\_cost}
\State Save and output $\mathbf{x}'$, the Boolean assignments of $\mathbf{x}$. Update best\_cost.
\EndIf
\State Calculate loss, $\mathcal{L}$, for $f(\mathbf{x})$ w.r.t variables contributing to the unsatisfied clauses, $\mathcal{U}$.
\State Run backpropagation. Run a single step of optimizer.
\EndWhile
\end{algorithmic}
\end{algorithm}

\textbf{Setup.}
We use PyTorch (v1.10.2) \citep{paszke2019pytorch}.
No other dependencies are required to run torchmSAT (e.g. no calls to a SAT oracle is made).
We use Adam optimizer for backprogation with a learning rate of\nobreakspace 1e-4.
For reproducibility and extensibility of our work, we use the PySAT toolkit (v0.1.8.dev1) \citep{imms-sat18} to synthesize hard problem instances and compare against existing methods.
The toolkit provides a unified interface to various state-of-the-art SAT and MaxSAT solvers.
The experimental results are obtained using a machine with Intel Xeon E5-2680 2x14cores@2.4 GHz, 128GB RAM.
For experiments on hardware acceleration, a Tesla P40 GPU is utilized to run torchmSAT.

\begin{figure}[t!]
\renewcommand{\tabcolsep}{0pt}
\vspace{-0.2in}
\begin{tabular}{cccc}
& \multicolumn{3}{c}{\includegraphics[clip, scale=0.35, valign=m]{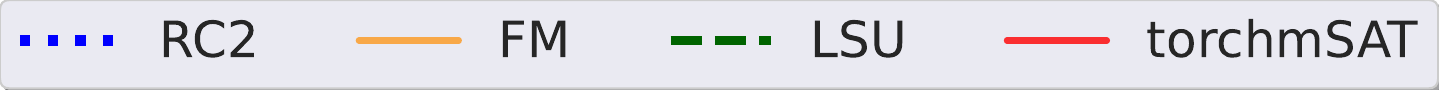}} \vspace{0.05in} \\
& T=1min. & T=5mins. & T=10mins. \\
\rotatebox[origin=c]{90}{\space\space\space\space\space\space\space\space\space\space CB} &
\includegraphics[clip, width=1.85in, height=1.4in, valign=m]{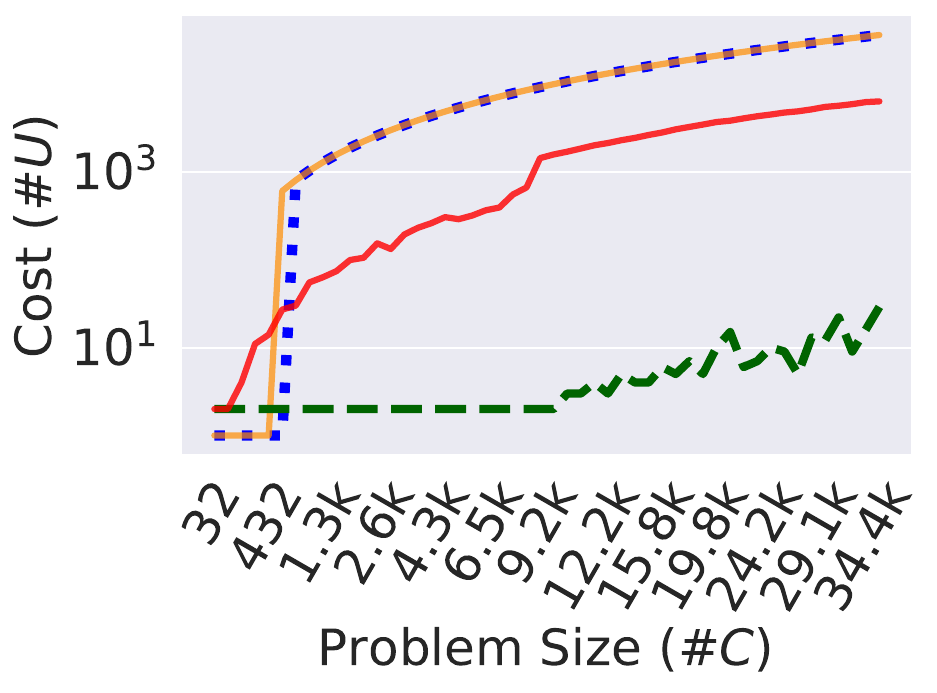} &
\hspace*{-0.3cm}
\includegraphics[clip, width=1.75in, height=1.4in, valign=m]{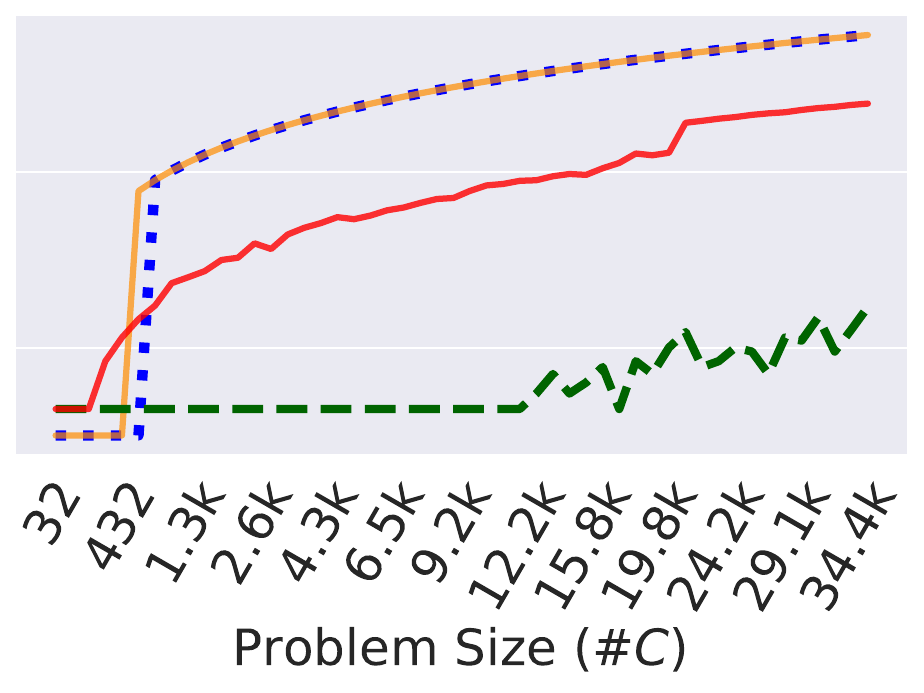}
 &
 \hspace*{-0.2cm}
\includegraphics[clip, width=1.75in, height=1.4in, valign=m]{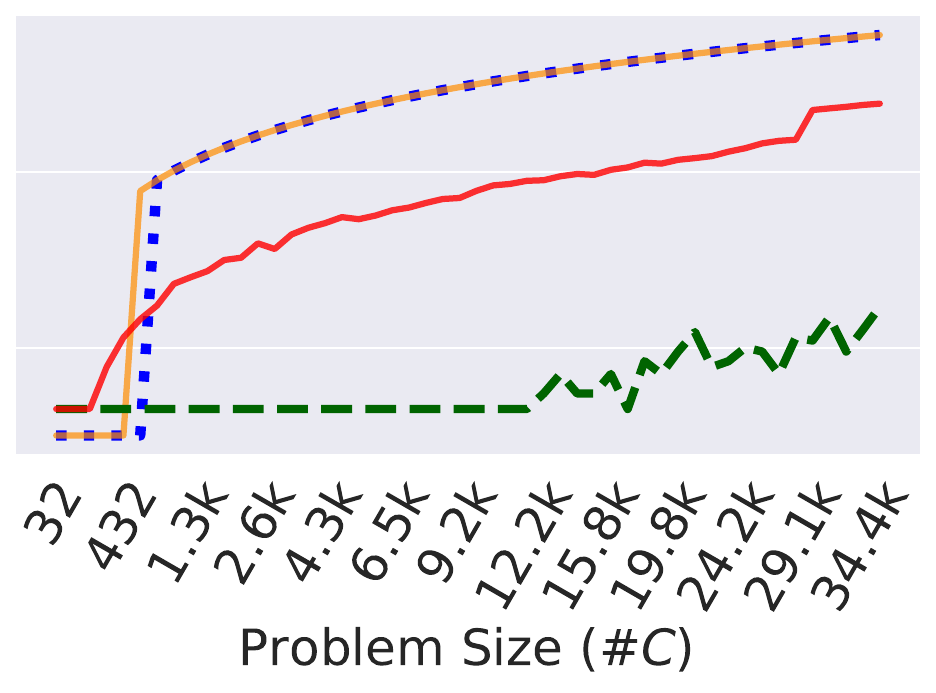}
\\
\rotatebox[origin=c]{90}
{\space\space\space\space\space\space\space\space\space\space\space\space GT} &
\includegraphics[clip, width=1.9in, height=1.4in, valign=m]{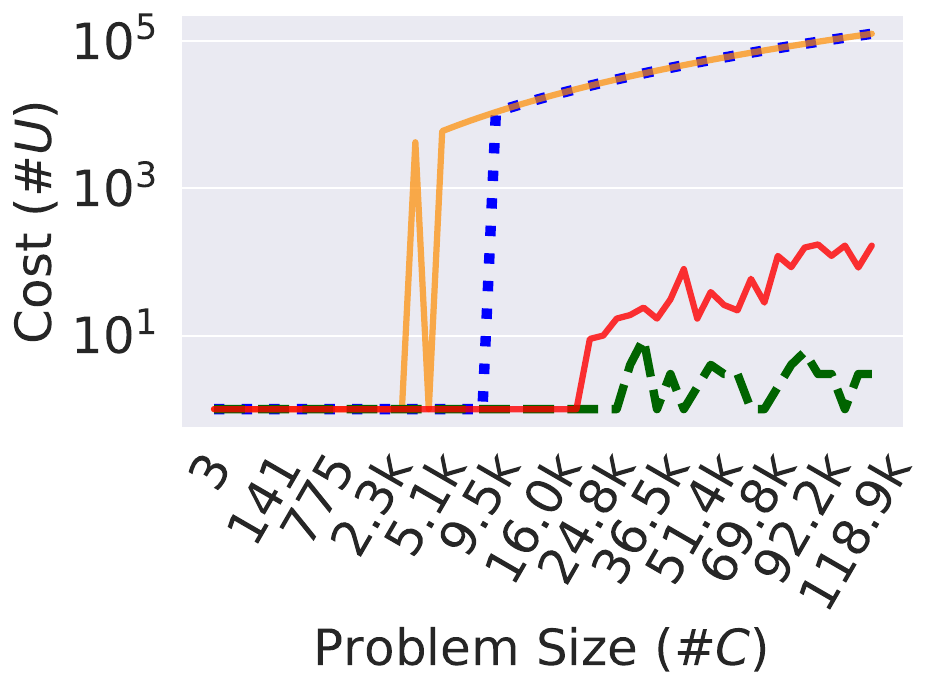} &
\hspace*{-0.3cm}
\includegraphics[clip, width=1.75in, height=1.4in, valign=m]{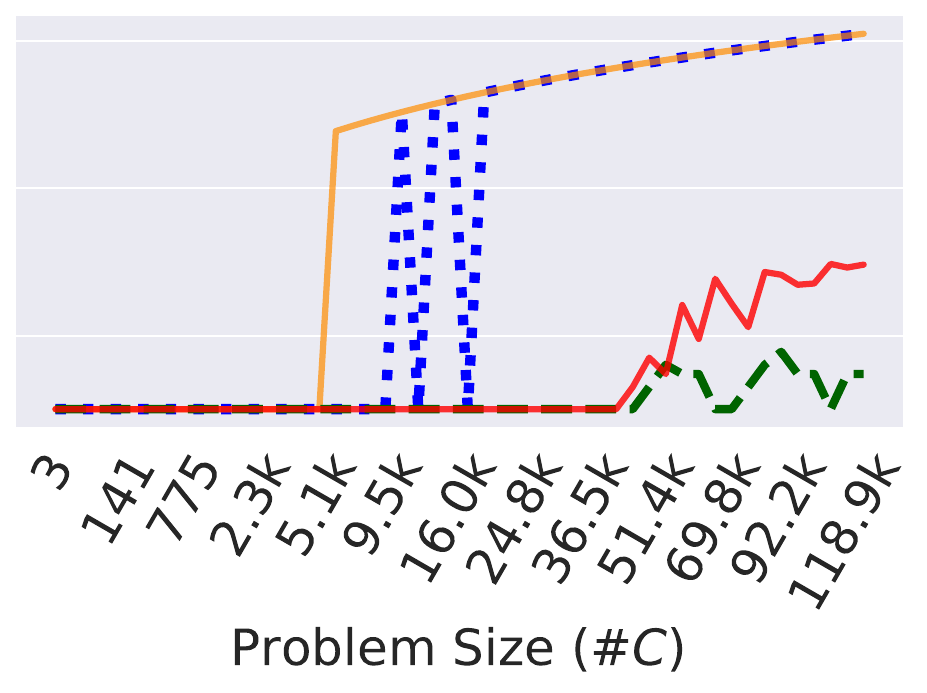} &
\hspace*{-0.3cm}
\includegraphics[clip, width=1.75in, height=1.4in, valign=m]{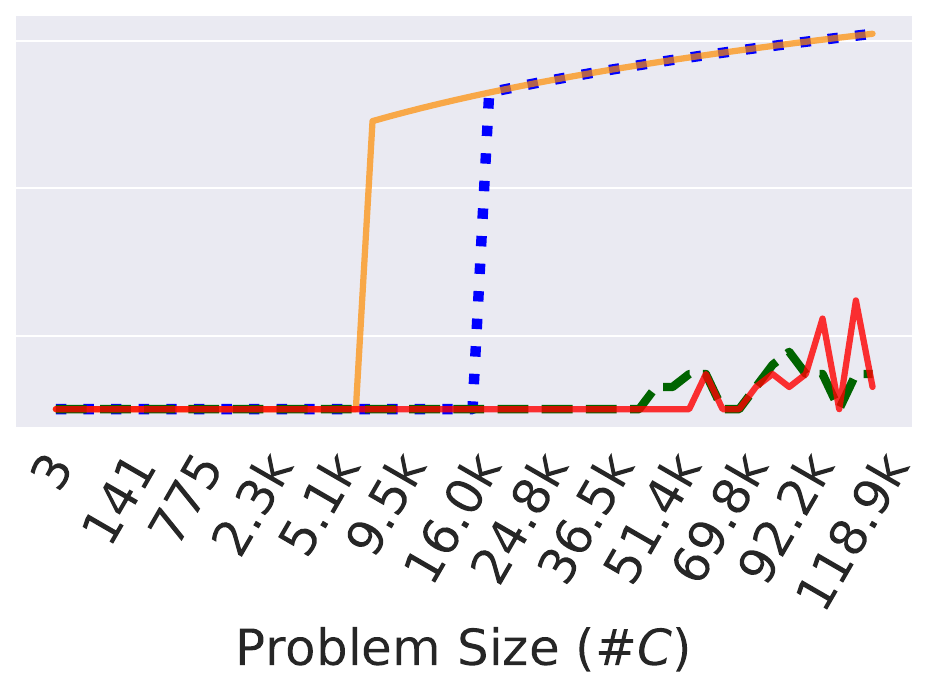}
\\

\rotatebox[origin=c]{90}{\space\space\space\space\space\space\space\space\space\space\space\space PAR} &
\includegraphics[clip, width=1.85in, height=1.4in, valign=m]{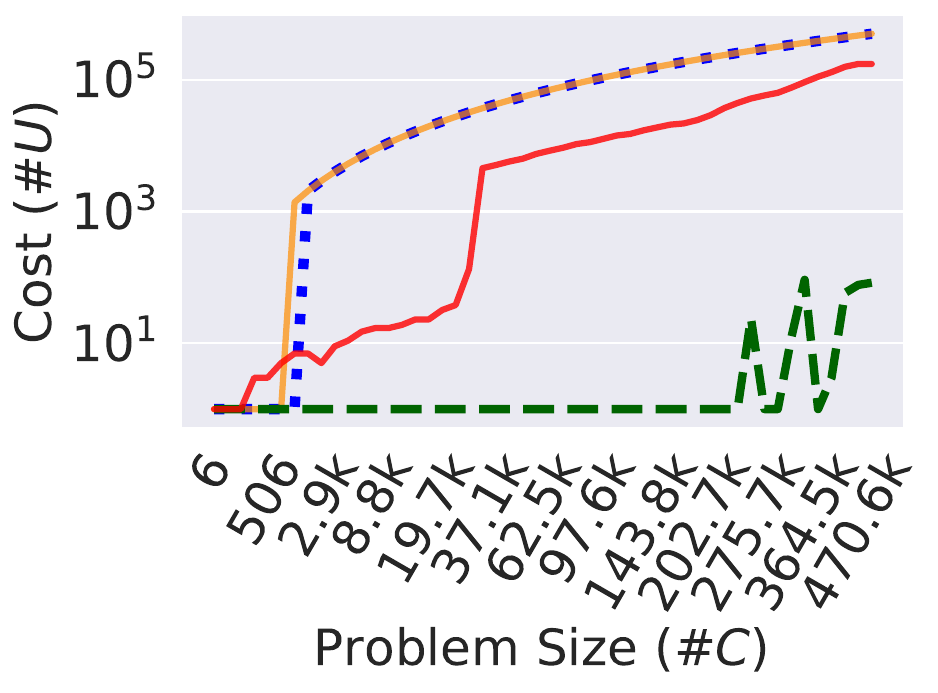} &
\hspace*{-0.4cm}
\includegraphics[clip, width=1.75in, height=1.4in, valign=m]{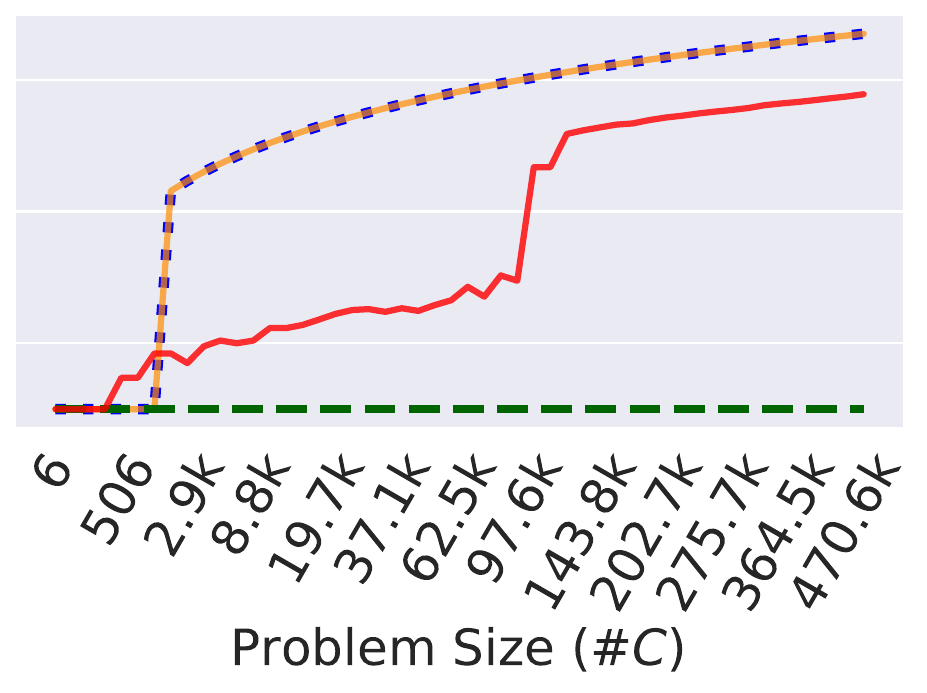} &
\hspace*{-0.4cm}
\includegraphics[clip, width=1.75in, height=1.4in, valign=m]{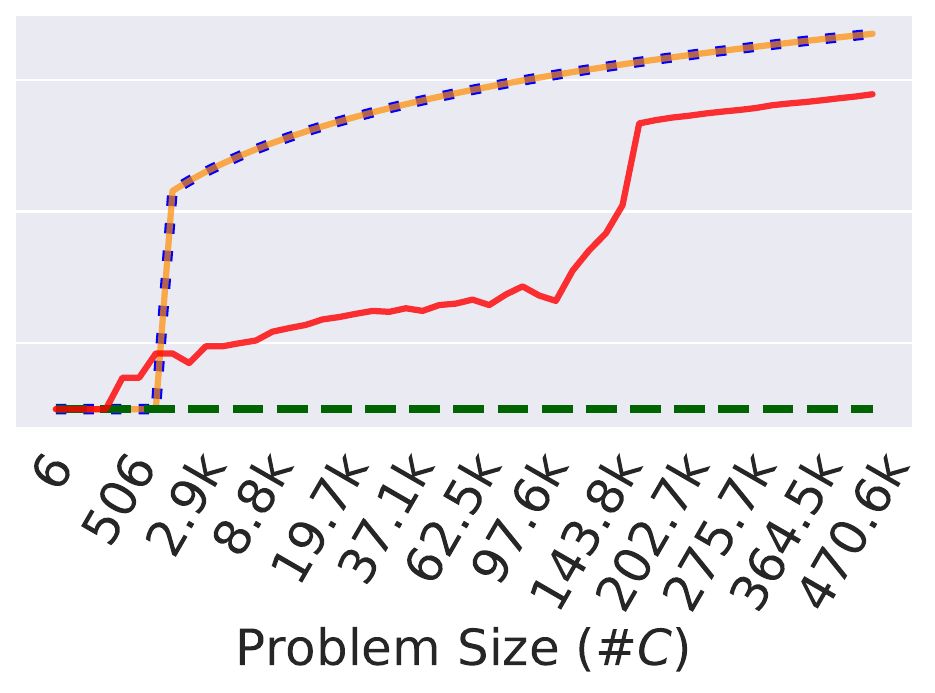}
\\

\rotatebox[origin=c]{90}{\space\space\space\space\space\space\space\space\space\space\space PHP} &
\hspace*{0.05cm}
\includegraphics[clip, width=1.85in, height=1.4in, valign=m]{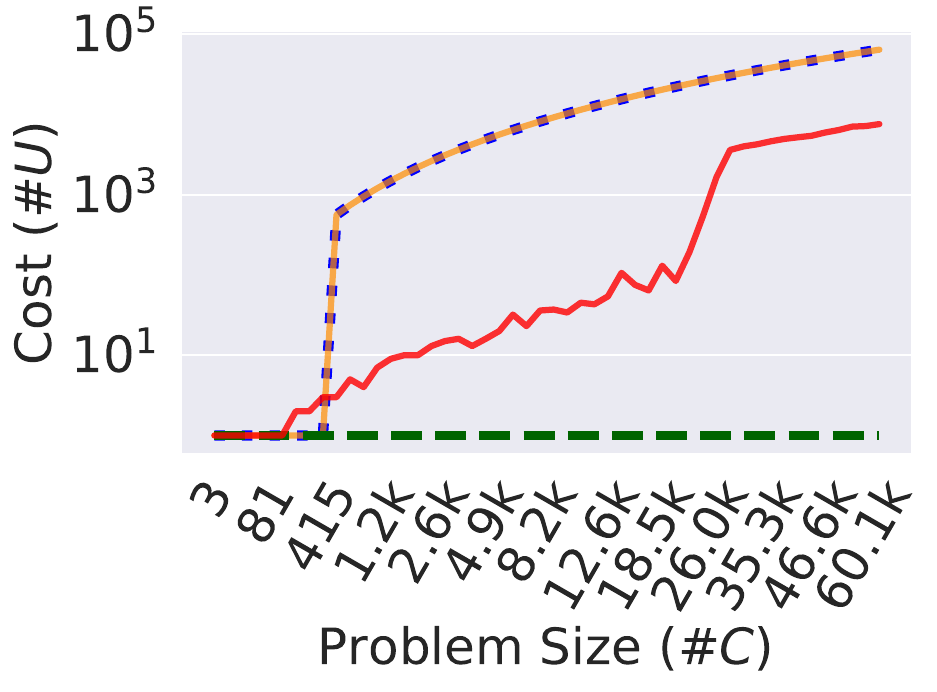} &
\hspace*{-0.2cm}
\includegraphics[clip, width=1.75in, height=1.4in, valign=m]{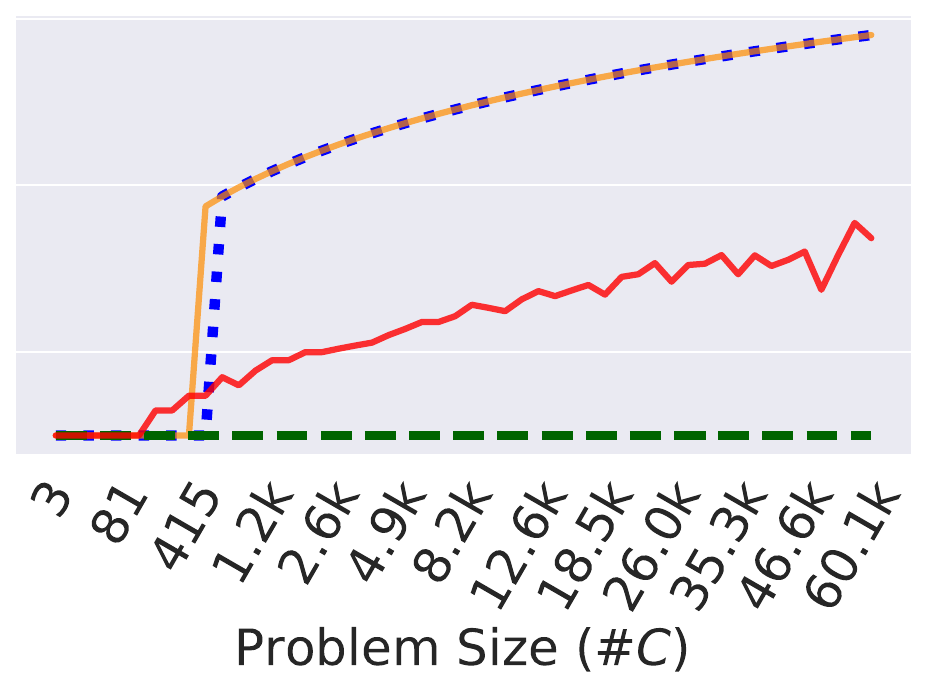} &
\hspace*{-0.2cm}
\includegraphics[clip, width=1.75in, height=1.4in, valign=m]{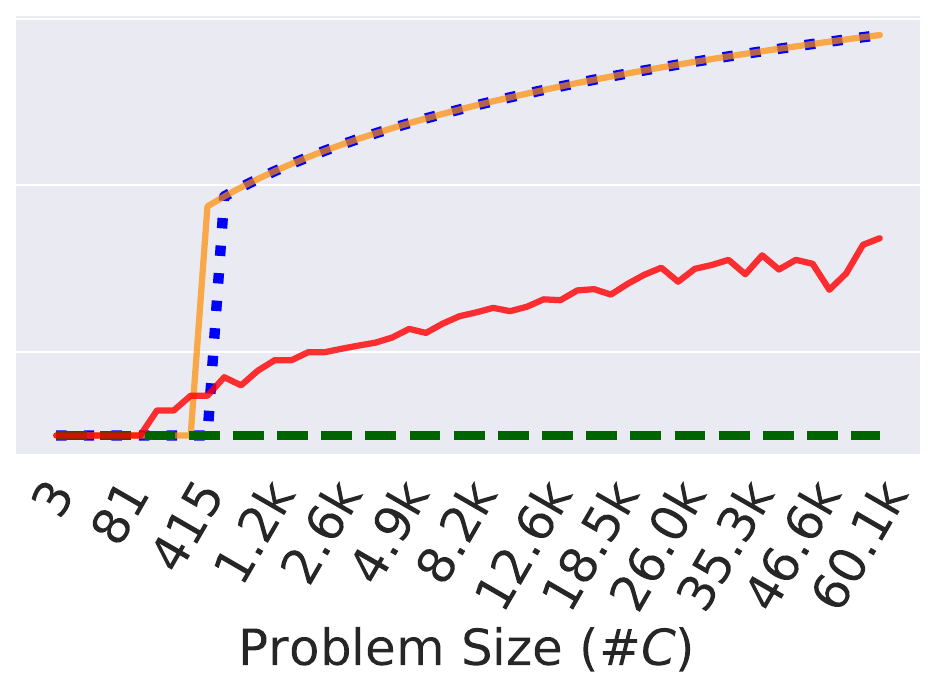}
\end{tabular}
\caption{Performance of torchmSAT as compared to the state-of-the-art MaxSAT solvers in PySAT \citep{imms-sat18}, namely FM, RC2 and LSU (refer to Section \ref{sec:preliminaries} for detailed descriptions of algorithms). Each row represents one of the datasets and each column is the time limit given to the solver. Each dataset contains 50 problem instances of increasing size (number of Boolean variables and number of clauses). In the plots, the problem size is parameterized by its total number of clauses on the x-axis. On the y-axis, the cost represents the number of unsatisfiable clauses ($\mathcal{U}$) by the end of the time limit. The FM and RC2 solvers tend to not perform well as problem instances get larger. The reason is that their solving process starts by calling the SAT oracle to establish the unsatisfiability of the given formula, which is NP-hard. LSU is the best-performing algorithm since it solves a given problem incrementally, delaying expensive calls to the SAT solver to the end. Our proposed approach, torchmSAT, is designed to be progressive like LSU, but without calling a SAT oracle. As evident from the plots, given more time, it becomes increasingly effective at finding solutions with lower costs. Synthesis scripts for the datasets are available in Appendix \ref{appendix:dataset}. PySAT solvers are called using their default configurations.}
\label{fig:comparison}
\vspace{-0.3in}
\end{figure}

\textbf{Evaluation Criteria.}
We will use a multi-faceted approach to thoroughly assess the performance of our proposed method.
First, we will investigate the cost obtained by our solver on problem instances of varying complexity and sizes, and under different time constraints.
The imposition of a time limit in this scenario is crucial given the NP-hard nature of MaxSAT, which implies that exploring and assessing the entire feasible region would necessitate exponential time.
Secondly, we will calculate the MaxSAT regret which quantifies the difference between the best solution found by any baseline solver and the solution obtained by our solver.
Lastly, we will assess the impact of leveraging GPU acceleration on the performance of our solver.

\textbf{Dataset.}
\label{sec:dataset}
While the datasets of the popular MaxSAT evaluations \footnote{\url{https://maxsat-evaluations.github.io/}} represent a set of non-trivial SAT instances, we opt to test our methods on smaller, yet hard, dataset to validate the applicability of the method.
Although our method does not outperform one of the state-of-the-art SAT solvers, torchmSAT offers a completely revamped method to solve MaxSAT problems without the need for a SAT oracle.
We employ PySAT to synthesize four representative datasets, encompassing combinatorial principles extensively examined in the context of propositional proof complexity.
Specifically, they implement encodings for the pigeonhole principle (PHP) \citep{cook1979relative}, the greater-than (ordering) principle (GT) \citep{krishnamurthy1985short}, the mutilated chessboard principle (CB) \citep{alekhnovich2004mutilated}, and the parity principle (PAR) \citep{ajtai1990parity}.
For each principle, we synthesize 50 MaxSAT instances of increasing sizes where all problem instances are not satisfiable.
In other words, there is at least one clause that cannot be satisfied, and the goal of MaxSAT solvers to find a feasible variable assignment that maximizes satisfiability.
Synthesis scripts are available in Appendix \ref{appendix:dataset}.

\textbf{Comparing with Existing MaxSAT Solvers.}
As discussed in Section \ref{sec:preliminaries}, existing MaxSAT algorithms proceed by making several calls to an underlying SAT oracle (i.e. solver).
In PySAT, both the RC2 and the FM algorithms start their solving process by making a call to the underlying SAT solver, which reports that $\mathcal{F}$ is unsatisfiable and returns an unsatisfiable core, $\mathcal{U}$.
The algorithm proceeds by alternating between relaxing clauses and calling the underlying SAT solver.
As depicted in Figure\nobreakspace \ref{fig:comparison}, both methods struggle to find any assignment for moderately large problem instances.
This is primarily due to their initial step, which involves invoking the SAT solver to establish unsatisfiability.
Conversely, the LSU algorithm identifies an initial assignment carrying a suboptimal cost, then incrementally searches for superior assignments to lower the cost.
This explains its position as the top-performing algorithm on the dataset.
Our proposed method, torchmSAT, excels at progressively generating viable solutions that surpass those of FM and RC2.
Indeed, when considering the GT dataset, torchmSAT's performance approaches that of LSU.

In Table \ref{table:detailed-results}, we compute the average regret of RC2, FM, and torchmSAT relative to LSU, the top-performing MaxSAT solver.
For every problem instance, we calculate a solver's regret as the difference between the cost it achieves and the cost LSU achieves.
As the results clearly show, torchmSAT generally demonstrates a markedly lower average regret.
Raw results for individual problem instances can be found in Appendix \ref{appendix:raw-results}.

\begin{figure}
\vspace{-0.1in}
\renewcommand{\tabcolsep}{0pt}
\centering
\begin{tabular}{cc}
\includegraphics[clip, scale=0.25, valign=m]{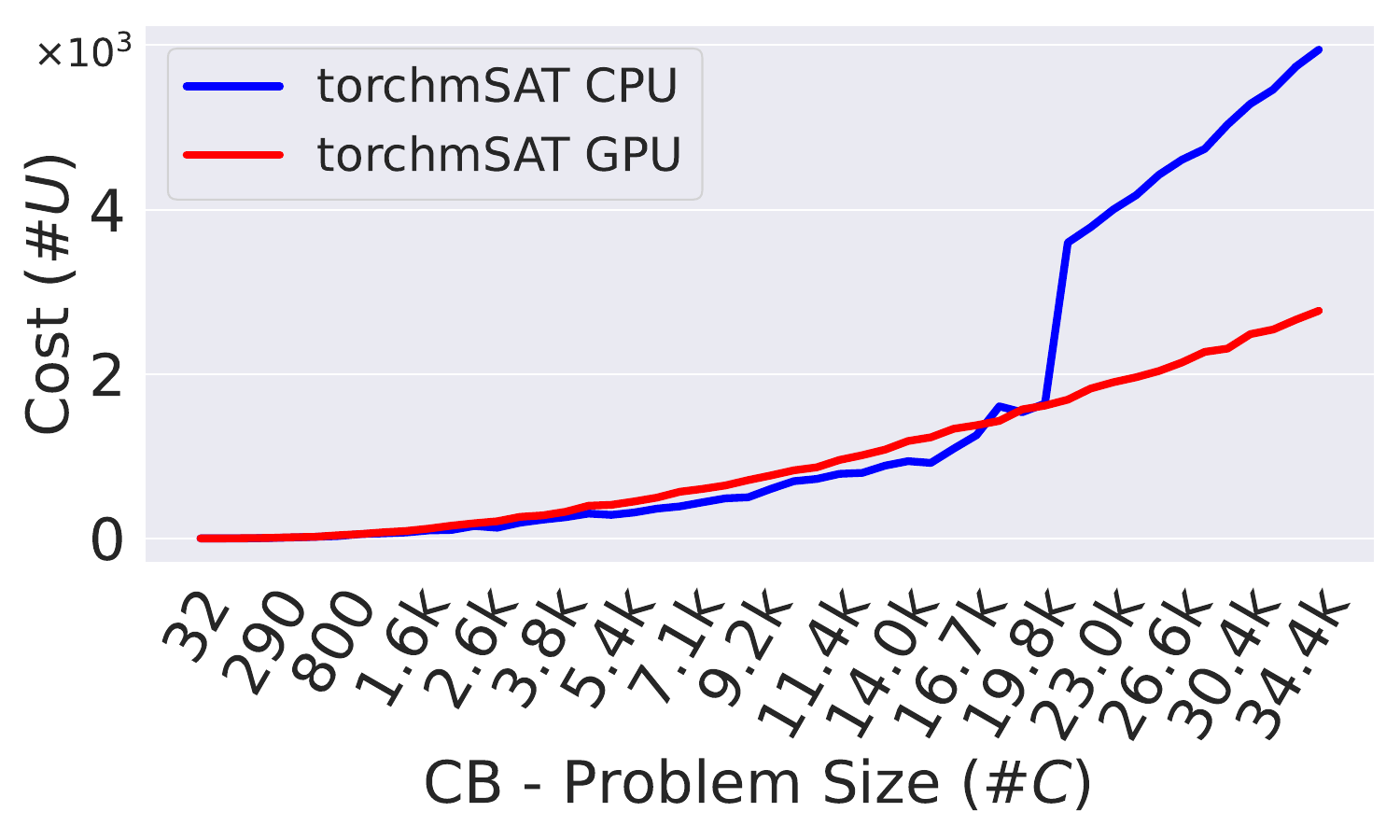} &
\includegraphics[clip, scale=0.25, valign=m]{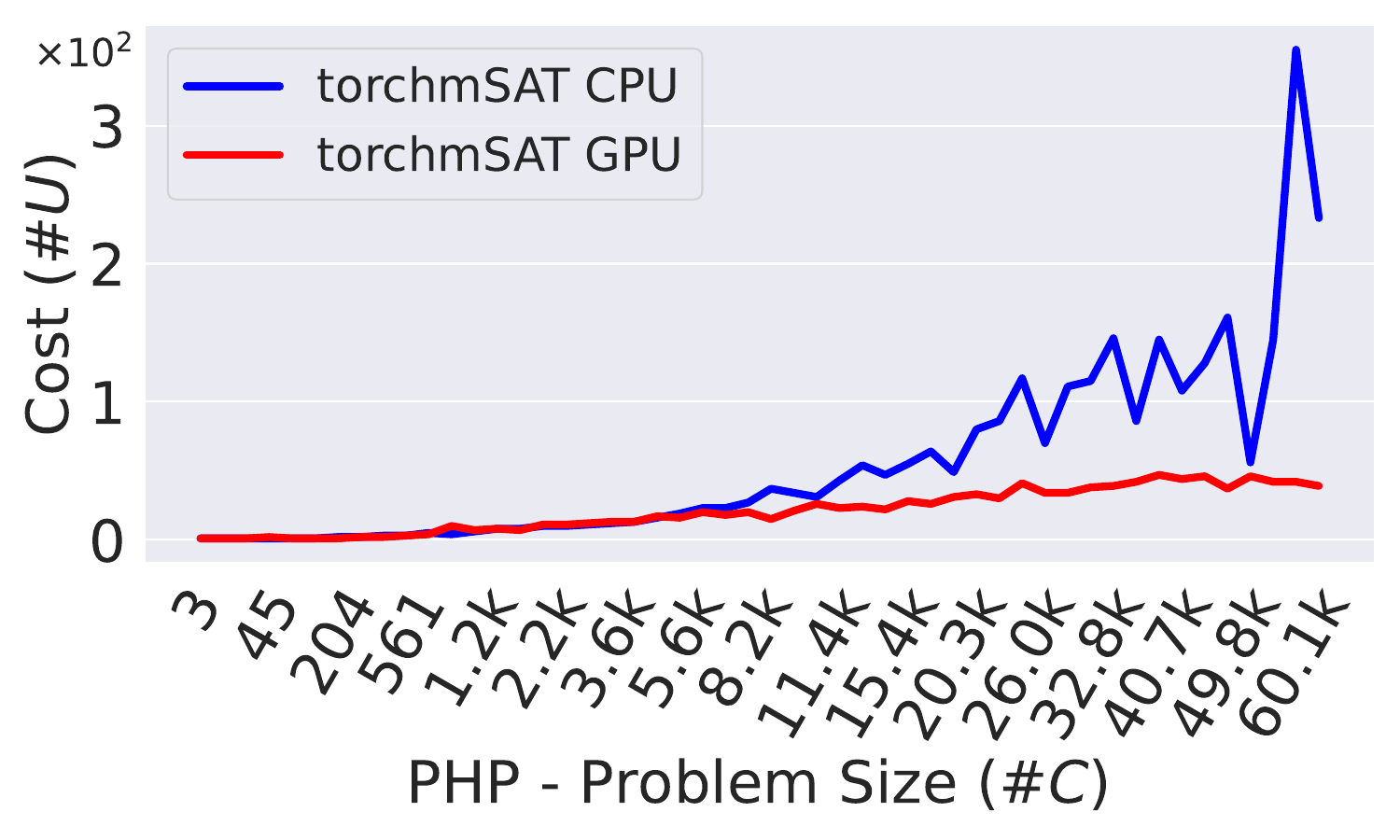}
\vspace{-0.1in}
\end{tabular}
\caption{Running torchmSAT on CPU vs. GPU, where it is capable of taking advantage of GPU acceleration, and finds better MaxSAT solutions within the same time limit (5mins). For complete results on scalability under different time limits, and on GT and PAR datasets, see Appendix \ref{appendix:gpu}.}
\label{fig:gpu-results}
\vspace{-0.2in}
\end{figure}

\textbf{GPU Acceleration.}
The main premise of our method is the novel application of contemporary GPUs and the evolving ecosystem of deep learning libraries and accelerators in solving MaxSAT instances.
In this section, we aim to demonstrate that by merely altering where the neural network is initialized, without any modifications to its structure or the solving process described earlier.
As shown in Figure\nobreakspace \ref{fig:gpu-results}, executing torchmSAT on a GPU yields solutions to larger MaxSAT instances with lower costs within the same time limit. 
This can be attributed to the enhanced speed of GPU computations, which accelerates the forward-loss-backward loop (Algorithm \ref{algo1}), thereby enabling more extensive exploration of the feasible region of variable assignments.

The use of GPUs in torchmSAT advances MaxSAT solving, offering an accelerated and efficient exploration of the solution space.
This acceleration is particularly transformative given that the algorithm's progressive nature means it benefits directly from more rapid computations, enabling it to find better solutions within the same time frame.
In contrast, traditional MaxSAT solvers are inherently sequential and cannot take advantage of GPU acceleration.

\section{Discussion}
\label{sec:conclusions}
\begin{table}[t]
  \caption{The average regret of the solvers. The regret($s$, $i$) of a solver $s$ on instance\nobreakspace $i$ is the difference between the cost of the best solution found by $s$ and the cost of best known solution:
regret($s$,\nobreakspace$i$) =  $cost_{s,i}\nobreakspace-\nobreakspace cost_{best, i}$. Cost is defined as the number of unsatisfied clauses, $\mathcal{U}$. Considering LSU is the the optimal solver (i.e. regret = 0), the table presents the average regret of torchmSAT as compared to FM and RC2. In torchmSAT, regret decreases as the solving time limit increases. See Appendix \ref{appendix:raw-results}.}
  \label{table:detailed-results}
  \footnotesize
  \centering
  \begin{tabular}{lrrr|rrr|rrr}
    \toprule
 & \multicolumn{3}{c}{RC2} & \multicolumn{3}{c}{FM} & \multicolumn{3}{c}{torchmSAT}         \\
    \cmidrule(r){2-10}
    Dataset & 1min & 5mins & 10mins & 1min & 5mins & 10mins & 1min  & 5mins & 10mins \\
    \midrule
    CB  & 12396 & 12396 & 12396 & 12408 & 12408 & 12408 & \textbf{1981} & \textbf{1496} & \textbf{1148}\\
    GT  & 31852 & 31245 & 30425 & 32552 & 32467 & 32206 & \textbf{30} & \textbf{13} & \textbf{2}\\
    PAR & 130013 & 130013 & 130013 & 130040 & 130013 & 130013 & \textbf{28272} & \textbf{13363} & \textbf{11777}\\
    PHP & 16688 & 16676 & 16676 & 16688 & 16688 & 16688 & \textbf{1384} & \textbf{55} & \textbf{47}\\
    \bottomrule
  \end{tabular}
\end{table}

\textbf{Limitations.}
While our proposed method presents several notable advantages, it also has some limitations.
First, it currently only works for unweighted MaxSAT instances, meaning it cannot handle instances where different clauses have different weights or importance.
Secondly, the algorithm lacks a definitive stopping criterion for backpropagation unless the optimal solution is zero.
This means it can be difficult to determine when the algorithm has reached an optimal solution or when to halt the process except using timeouts.
Lastly, the memory requirement for the method is $\mathcal{O}(nm)$, which can be prohibitive for large instances.
Although the matrix W is sparse, which could potentially be leveraged to save memory, our current implementation does not take advantage of this sparsity. 

\textbf{Conclusions and Future Work.}
In conclusion, we have presented a new method for MaxSAT solving that capitalizes on the use of neural networks.
Our method, named torchmSAT, is a progressive approach that continually refines and improves its solutions over time.
One of the core advantages of torchmSAT is its independence from a SAT oracle, a feature that differentiates it from traditional MaxSAT solvers.
This makes our method more self-sufficient and less reliant on external components.
Experimental results show that our method outperforms two existing MaxSAT solvers, and is on par with another state-of-the-art solver for small to medium problem sizes.
Additionally, torchmSAT is able to benefit from GPU acceleration, allowing for more rapid exploration of feasible solution regions.
Despite some limitations, torchmSAT represents a promising step forward in MaxSAT problem solving. 
For future work, we aim to extend the capabilities of our method to handle weighted MaxSAT instances, develop a clear stopping criteria for backpropagation, and optimize memory usage by leveraging the sparsity of the matrix $W$.
This could lead to a more efficient, versatile and powerful solver that can tackle even more complex problems.

\textbf{Reproducibility.}
The implementation of torchmSAT is provided as a Python package in the supplementary material.
The dataset can be synthesized using scripts in Appendix \ref{appendix:dataset}.

\bibliographystyle{plainnat}
\bibliography{refbib}

\newpage
\section*{Appendices}

\appendix

\section{Dataset Details}
\label{appendix:dataset}
We employ PySAT to synthesize four representative datasets, encompassing combinatorial principles extensively examined in the context of propositional proof complexity.
Specifically, they implement encodings for the pigeonhole principle (PHP) \citep{cook1979relative}, the greater-than (ordering) principle (GT) \citep{krishnamurthy1985short}, the mutilated chessboard principle (CB) \citep{alekhnovich2004mutilated}, and the parity principle (PAR) \citep{ajtai1990parity}.
For each principle, we synthesize 50 MaxSAT instances of increasing sizes where all problem instances are not satisfiable.
In other words, there is at least one clause that cannot be satisfied, and the goal of MaxSAT solvers to find a feasible variable assignment that maximizes satisfiability.

Below is the synthesis script.

\begin{lstlisting}[language=Python, caption=Synthesis script for the datasets, label=codesample]
from pysat.examples.genhard import CB, GT, PAR, PHP


def gen_php():
    for n_holes in range(1, 51):
        cnf = PHP(n_holes)
        cnf.to_file(fname=f"data/php/{cnf.nv}_{len(cnf.clauses)}_{n_holes}.zip")

def gen_cb():
    for size in range(1, 51):
        cnf = CB(size)
        cnf.to_file(fname=f"data/cb/{cnf.nv}_{len(cnf.clauses)}_{size}.zip")

def gen_gt():
    for size in range(1, 51):
        cnf = GT(size)
        cnf.to_file(fname=f"data/gt/{cnf.nv}_{len(cnf.clauses)}_{size}.zip")

def gen_par():
    for size in range(1, 51):
        cnf = PAR(size)
        cnf.to_file(fname=f"data/par/{cnf.nv}_{len(cnf.clauses)}_{size}.zip")
\end{lstlisting}

For each dataset, we provide detailed statistics on the number of Boolean variables and clauses in Table \ref{table:detailed-dataset-stats}.

\begin{table}[t!]
  \caption{Number of Boolean variables and clauses in each problem instance in the datasets. This indicates the difficulty level of solving each instance.}
  \label{table:detailed-dataset-stats}
  \centering
  \begin{tabular}{lrr|rr|rr|rr}
    \toprule
 & \multicolumn{2}{c}{\textbf{CB \citep{alekhnovich2004mutilated}}} & \multicolumn{2}{c}{\textbf{GT \citep{krishnamurthy1985short}}} & \multicolumn{2}{c}{\textbf{PAR \citep{ajtai1990parity}}} & \multicolumn{2}{c}{\textbf{PHP \citep{cook1979relative}}} \\
    \cmidrule(r){2-9}
    & \# vars & \# clauses & \# vars & \# clauses & \# vars & \# clauses & \# vars & \# clauses \\
    \midrule
    1 & 20 & 32 & 2 & 3 & 3 & 6 & 2 & 3 \\
2 & 56 & 90 & 6 & 12 & 10 & 35 & 6 & 9 \\
3 & 108 & 176 & 12 & 34 & 21 & 112 & 12 & 22 \\
4 & 176 & 290 & 20 & 75 & 36 & 261 & 20 & 45 \\
5 & 260 & 432 & 30 & 141 & 55 & 506 & 30 & 81 \\
6 & 360 & 602 & 42 & 238 & 78 & 871 & 42 & 133 \\
7 & 476 & 800 & 56 & 372 & 105 & 1380 & 56 & 204 \\
8 & 608 & 1026 & 72 & 549 & 136 & 2057 & 72 & 297 \\
9 & 756 & 1280 & 90 & 775 & 171 & 2926 & 90 & 415 \\
10 & 920 & 1562 & 110 & 1056 & 210 & 4011 & 110 & 561 \\
11 & 1100 & 1872 & 132 & 1398 & 253 & 5336 & 132 & 738 \\
12 & 1296 & 2210 & 156 & 1807 & 300 & 6925 & 156 & 949 \\
13 & 1508 & 2576 & 182 & 2289 & 351 & 8802 & 182 & 1197 \\
14 & 1736 & 2970 & 210 & 2850 & 406 & 10991 & 210 & 1485 \\
15 & 1980 & 3392 & 240 & 3496 & 465 & 13516 & 240 & 1816 \\
16 & 2240 & 3842 & 272 & 4233 & 528 & 16401 & 272 & 2193 \\
17 & 2516 & 4320 & 306 & 5067 & 595 & 19670 & 306 & 2619 \\
18 & 2808 & 4826 & 342 & 6004 & 666 & 23347 & 342 & 3097 \\
19 & 3116 & 5360 & 380 & 7050 & 741 & 27456 & 380 & 3630 \\
20 & 3440 & 5922 & 420 & 8211 & 820 & 32021 & 420 & 4221 \\
21 & 3780 & 6512 & 462 & 9493 & 903 & 37066 & 462 & 4873 \\
22 & 4136 & 7130 & 506 & 10902 & 990 & 42615 & 506 & 5589 \\
23 & 4508 & 7776 & 552 & 12444 & 1081 & 48692 & 552 & 6372 \\
24 & 4896 & 8450 & 600 & 14125 & 1176 & 55321 & 600 & 7225 \\
25 & 5300 & 9152 & 650 & 15951 & 1275 & 62526 & 650 & 8151 \\
26 & 5720 & 9882 & 702 & 17928 & 1378 & 70331 & 702 & 9153 \\
27 & 6156 & 10640 & 756 & 20062 & 1485 & 78760 & 756 & 10234 \\
28 & 6608 & 11426 & 812 & 22359 & 1596 & 87837 & 812 & 11397 \\
29 & 7076 & 12240 & 870 & 24825 & 1711 & 97586 & 870 & 12645 \\
30 & 7560 & 13082 & 930 & 27466 & 1830 & 108031 & 930 & 13981 \\
31 & 8060 & 13952 & 992 & 30288 & 1953 & 119196 & 992 & 15408 \\
32 & 8576 & 14850 & 1056 & 33297 & 2080 & 131105 & 1056 & 16929 \\
33 & 9108 & 15776 & 1122 & 36499 & 2211 & 143782 & 1122 & 18547 \\
34 & 9656 & 16730 & 1190 & 39900 & 2346 & 157251 & 1190 & 20265 \\
35 & 10220 & 17712 & 1260 & 43506 & 2485 & 171536 & 1260 & 22086 \\
36 & 10800 & 18722 & 1332 & 47323 & 2628 & 186661 & 1332 & 24013 \\
37 & 11396 & 19760 & 1406 & 51357 & 2775 & 202650 & 1406 & 26049 \\
38 & 12008 & 20826 & 1482 & 55614 & 2926 & 219527 & 1482 & 28197 \\
39 & 12636 & 21920 & 1560 & 60100 & 3081 & 237316 & 1560 & 30460 \\
40 & 13280 & 23042 & 1640 & 64821 & 3240 & 256041 & 1640 & 32841 \\
41 & 13940 & 24192 & 1722 & 69783 & 3403 & 275726 & 1722 & 35343 \\
42 & 14616 & 25370 & 1806 & 74992 & 3570 & 296395 & 1806 & 37969 \\
43 & 15308 & 26576 & 1892 & 80454 & 3741 & 318072 & 1892 & 40722 \\
44 & 16016 & 27810 & 1980 & 86175 & 3916 & 340781 & 1980 & 43605 \\
45 & 16740 & 29072 & 2070 & 92161 & 4095 & 364546 & 2070 & 46621 \\
46 & 17480 & 30362 & 2162 & 98418 & 4278 & 389391 & 2162 & 49773 \\
47 & 18236 & 31680 & 2256 & 104952 & 4465 & 415340 & 2256 & 53064 \\
48 & 19008 & 33026 & 2352 & 111769 & 4656 & 442417 & 2352 & 56497 \\
49 & 19796 & 34400 & 2450 & 118875 & 4851 & 470646 & 2450 & 60075 \\
50 & 20600 & 35802 & 2550 & 126276 & 5050 & 500051 & 2550 & 63801 \\
    \bottomrule
  \end{tabular}
\end{table}

\section{Raw Results}
\label{appendix:raw-results}
In Table \ref{table:raw-results}, we provide the raw results discussed in the Experiments Section \ref{sec:expr}.
It is a detailed expansion of Table \ref{table:detailed-results}.
The goal is to give a deeper look into Figure \ref{fig:comparison}, and see how numbers relate to each other.

\begin{table}[]
  \caption{Raw results for costs (i.e. number of unsatisfied clauses) obtained by different solvers.}
  \label{table:raw-results}
  \centering
  \footnotesize
  \begin{longtable}{llrrr|rrr|rrr|rrr}
    \toprule
 & & \multicolumn{3}{c}{RC2} & \multicolumn{3}{c}{FM} & \multicolumn{3}{c}{LSU} & \multicolumn{3}{c}{torchmSAT}\\
    \cmidrule(r){3-14}
    & & 1m & 5m & 10m & 1m & 5m & 10m & 1m  & 5m & 10m & 1m  & 5m & 10m \\
    \midrule
    \endhead
    
    CB & 1 & 1 & 1 & 1 & 1 & 1 & 1 & 2 & 2 & 2 & 2 & 2 & 2 \\
& 2 & 1 & 1 & 1 & 1 & 1 & 1 & 2 & 2 & 2 & 2 & 2 & 2 \\
& 3 & 1 & 1 & 1 & 1 & 1 & 1 & 2 & 2 & 2 & 4 & 2 & 2 \\
& 4 & 1 & 1 & 1 & 1 & 1 & 1 & 2 & 2 & 2 & 11 & 7 & 6 \\
& 5 & 1 & 1 & 1 & 1 & 1 & 1 & 2 & 2 & 2 & 14 & 13 & 13 \\
& 6 & 1 & 1 & 1 & 602 & 602 & 602 & 2 & 2 & 2 & 27 & 21 & 21 \\
& 7 & 800 & 800 & 800 & 800 & 800 & 800 & 2 & 2 & 2 & 30 & 30 & 30 \\
& 8 & 1026 & 1026 & 1026 & 1026 & 1026 & 1026 & 2 & 2 & 2 & 55 & 54 & 53 \\
& 9 & 1280 & 1280 & 1280 & 1280 & 1280 & 1280 & 2 & 2 & 2 & 63 & 63 & 63 \\
& 10 & 1562 & 1562 & 1562 & 1562 & 1562 & 1562 & 2 & 2 & 2 & 74 & 74 & 74 \\
& 11 & 1872 & 1872 & 1872 & 1872 & 1872 & 1872 & 2 & 2 & 2 & 99 & 99 & 99 \\
& 12 & 2210 & 2210 & 2210 & 2210 & 2210 & 2210 & 2 & 2 & 2 & 105 & 105 & 105 \\
& 13 & 2576 & 2576 & 2576 & 2576 & 2576 & 2576 & 2 & 2 & 2 & 153 & 153 & 153 \\
& 14 & 2970 & 2970 & 2970 & 2970 & 2970 & 2970 & 2 & 2 & 2 & 132 & 132 & 132 \\
& 15 & 3392 & 3392 & 3392 & 3392 & 3392 & 3392 & 2 & 2 & 2 & 193 & 193 & 193 \\
& 16 & 3842 & 3842 & 3842 & 3842 & 3842 & 3842 & 2 & 2 & 2 & 230 & 230 & 230 \\
& 17 & 4320 & 4320 & 4320 & 4320 & 4320 & 4320 & 2 & 2 & 2 & 260 & 260 & 260 \\
& 18 & 4826 & 4826 & 4826 & 4826 & 4826 & 4826 & 2 & 2 & 2 & 304 & 304 & 304 \\
& 19 & 5360 & 5360 & 5360 & 5360 & 5360 & 5360 & 2 & 2 & 2 & 288 & 288 & 288 \\
& 20 & 5922 & 5922 & 5922 & 5922 & 5922 & 5922 & 2 & 2 & 2 & 317 & 317 & 317 \\
& 21 & 6512 & 6512 & 6512 & 6512 & 6512 & 6512 & 2 & 2 & 2 & 364 & 364 & 364 \\
& 22 & 7130 & 7130 & 7130 & 7130 & 7130 & 7130 & 2 & 2 & 2 & 391 & 391 & 391 \\
& 23 & 7776 & 7776 & 7776 & 7776 & 7776 & 7776 & 2 & 2 & 2 & 550 & 441 & 441 \\
& 24 & 8450 & 8450 & 8450 & 8450 & 8450 & 8450 & 2 & 2 & 2 & 661 & 489 & 489 \\
& 25 & 9152 & 9152 & 9152 & 9152 & 9152 & 9152 & 2 & 2 & 2 & 1426 & 502 & 502 \\
& 26 & 9882 & 9882 & 9882 & 9882 & 9882 & 9882 & 2 & 2 & 2 & 1569 & 605 & 605 \\
& 27 & 10640 & 10640 & 10640 & 10640 & 10640 & 10640 & 3 & 2 & 2 & 1687 & 698 & 698 \\
& 28 & 11426 & 11426 & 11426 & 11426 & 11426 & 11426 & 3 & 2 & 2 & 1833 & 725 & 725 \\
& 29 & 12240 & 12240 & 12240 & 12240 & 12240 & 12240 & 4 & 2 & 2 & 1995 & 787 & 787 \\
& 30 & 13082 & 13082 & 13082 & 13082 & 13082 & 13082 & 3 & 3 & 3 & 2116 & 799 & 799 \\
& 31 & 13952 & 13952 & 13952 & 13952 & 13952 & 13952 & 5 & 5 & 5 & 2280 & 887 & 887 \\
& 32 & 14850 & 14850 & 14850 & 14850 & 14850 & 14850 & 4 & 3 & 3 & 2423 & 941 & 941 \\
& 33 & 15776 & 15776 & 15776 & 15776 & 15776 & 15776 & 4 & 4 & 3 & 2614 & 920 & 920 \\
& 34 & 16730 & 16730 & 16730 & 16730 & 16730 & 16730 & 6 & 6 & 5 & 2790 & 1093 & 1050 \\
& 35 & 17712 & 17712 & 17712 & 17712 & 17712 & 17712 & 5 & 2 & 2 & 3030 & 1257 & 1117 \\
& 36 & 18722 & 18722 & 18722 & 18722 & 18722 & 18722 & 7 & 7 & 7 & 3227 & 1608 & 1265 \\
& 37 & 19760 & 19760 & 19760 & 19760 & 19760 & 19760 & 5 & 5 & 5 & 3433 & 1536 & 1234 \\
& 38 & 20826 & 20826 & 20826 & 20826 & 20826 & 20826 & 10 & 10 & 9 & 3669 & 1641 & 1364 \\
& 39 & 21920 & 21920 & 21920 & 21920 & 21920 & 21920 & 15 & 15 & 15 & 3797 & 3599 & 1426 \\
& 40 & 23042 & 23042 & 23042 & 23042 & 23042 & 23042 & 6 & 6 & 6 & 4036 & 3785 & 1502 \\
& 41 & 24192 & 24192 & 24192 & 24192 & 24192 & 24192 & 7 & 7 & 7 & 4268 & 4002 & 1688 \\
& 42 & 25370 & 25370 & 25370 & 25370 & 25370 & 25370 & 10 & 10 & 10 & 4465 & 4176 & 1849 \\
& 43 & 26576 & 26576 & 26576 & 26576 & 26576 & 26576 & 9 & 9 & 9 & 4701 & 4424 & 2097 \\
& 44 & 27810 & 27810 & 27810 & 27810 & 27810 & 27810 & 5 & 5 & 5 & 4855 & 4606 & 2242 \\
& 45 & 29072 & 29072 & 29072 & 29072 & 29072 & 29072 & 13 & 13 & 13 & 5109 & 4738 & 2307 \\
& 46 & 30362 & 30362 & 30362 & 30362 & 30362 & 30362 & 12 & 12 & 12 & 5449 & 5034 & 5025 \\
& 47 & 31680 & 31680 & 31680 & 31680 & 31680 & 31680 & 22 & 22 & 22 & 5621 & 5286 & 5243 \\
& 48 & 33026 & 33026 & 33026 & 33026 & 33026 & 33026 & 9 & 9 & 9 & 5857 & 5460 & 5460 \\
& 49 & 34400 & 34400 & 34400 & 34400 & 34400 & 34400 & 16 & 16 & 16 & 6182 & 5740 & 5736 \\
& 50 & 35802 & 35802 & 35802 & 35802 & 35802 & 35802 & 29 & 29 & 29 & 6297 & 5945 & 5936 \\
    \midrule
    \hline \multicolumn{14}{|r|}{{Continued on next page}} \\ \hline
  \end{longtable}
\end{table}

\begin{table}[]
  \centering
  \footnotesize
  \begin{longtable}{llrrr|rrr|rrr|rrr}
    \toprule
 & & \multicolumn{3}{c}{RC2} & \multicolumn{3}{c}{FM} & \multicolumn{3}{c}{LSU} & \multicolumn{3}{c}{torchmSAT}\\
    \cmidrule(r){3-14}
    & & 1m & 5m & 10m & 1m & 5m & 10m & 1m  & 5m & 10m & 1m  & 5m & 10m \\
    \midrule
    \endhead
    GT & 1 & 1 & 1 & 1 & 1 & 1 & 1 & 1 & 1 & 1 & 1 & 1 & 1 \\
& 2 & 1 & 1 & 1 & 1 & 1 & 1 & 1 & 1 & 1 & 1 & 1 & 1 \\
& 3 & 1 & 1 & 1 & 1 & 1 & 1 & 1 & 1 & 1 & 1 & 1 & 1 \\
& 4 & 1 & 1 & 1 & 1 & 1 & 1 & 1 & 1 & 1 & 1 & 1 & 1 \\
& 5 & 1 & 1 & 1 & 1 & 1 & 1 & 1 & 1 & 1 & 1 & 1 & 1 \\
& 6 & 1 & 1 & 1 & 1 & 1 & 1 & 1 & 1 & 1 & 1 & 1 & 1 \\
& 7 & 1 & 1 & 1 & 1 & 1 & 1 & 1 & 1 & 1 & 1 & 1 & 1 \\
& 8 & 1 & 1 & 1 & 1 & 1 & 1 & 1 & 1 & 1 & 1 & 1 & 1 \\
& 9 & 1 & 1 & 1 & 1 & 1 & 1 & 1 & 1 & 1 & 1 & 1 & 1 \\
& 10 & 1 & 1 & 1 & 1 & 1 & 1 & 1 & 1 & 1 & 1 & 1 & 1 \\
& 11 & 1 & 1 & 1 & 1 & 1 & 1 & 1 & 1 & 1 & 1 & 1 & 1 \\
& 12 & 1 & 1 & 1 & 1 & 1 & 1 & 1 & 1 & 1 & 1 & 1 & 1 \\
& 13 & 1 & 1 & 1 & 1 & 1 & 1 & 1 & 1 & 1 & 1 & 1 & 1 \\
& 14 & 1 & 1 & 1 & 1 & 1 & 1 & 1 & 1 & 1 & 1 & 1 & 1 \\
& 15 & 1 & 1 & 1 & 1 & 1 & 1 & 1 & 1 & 1 & 1 & 1 & 1 \\
& 16 & 1 & 1 & 1 & 4233 & 1 & 1 & 1 & 1 & 1 & 1 & 1 & 1 \\
& 17 & 1 & 1 & 1 & 1 & 1 & 1 & 1 & 1 & 1 & 1 & 1 & 1 \\
& 18 & 1 & 1 & 1 & 6004 & 6004 & 1 & 1 & 1 & 1 & 1 & 1 & 1 \\
& 19 & 1 & 1 & 1 & 7050 & 7050 & 1 & 1 & 1 & 1 & 1 & 1 & 1 \\
& 20 & 1 & 1 & 1 & 8211 & 8211 & 8211 & 1 & 1 & 1 & 1 & 1 & 1 \\
& 21 & 1 & 1 & 1 & 9493 & 9493 & 9493 & 1 & 1 & 1 & 1 & 1 & 1 \\
& 22 & 10902 & 10902 & 1 & 10902 & 10902 & 10902 & 1 & 1 & 1 & 1 & 1 & 1 \\
& 23 & 12444 & 1 & 1 & 12444 & 12444 & 12444 & 1 & 1 & 1 & 1 & 1 & 1 \\
& 24 & 14125 & 14125 & 1 & 14125 & 14125 & 14125 & 1 & 1 & 1 & 1 & 1 & 1 \\
& 25 & 15951 & 15951 & 1 & 15951 & 15951 & 15951 & 1 & 1 & 1 & 1 & 1 & 1 \\
& 26 & 17928 & 1 & 1 & 17928 & 17928 & 17928 & 1 & 1 & 1 & 1 & 1 & 1 \\
& 27 & 20062 & 20062 & 20062 & 20062 & 20062 & 20062 & 1 & 1 & 1 & 1 & 1 & 1 \\
& 28 & 22359 & 22359 & 22359 & 22359 & 22359 & 22359 & 1 & 1 & 1 & 1 & 1 & 1 \\
& 29 & 24825 & 24825 & 24825 & 24825 & 24825 & 24825 & 1 & 1 & 1 & 9 & 1 & 1 \\
& 30 & 27466 & 27466 & 27466 & 27466 & 27466 & 27466 & 1 & 1 & 1 & 10 & 1 & 1 \\
& 31 & 30288 & 30288 & 30288 & 30288 & 30288 & 30288 & 1 & 1 & 1 & 17 & 1 & 1 \\
& 32 & 33297 & 33297 & 33297 & 33297 & 33297 & 33297 & 4 & 1 & 1 & 19 & 1 & 1 \\
& 33 & 36499 & 36499 & 36499 & 36499 & 36499 & 36499 & 9 & 1 & 1 & 24 & 1 & 1 \\
& 34 & 39900 & 39900 & 39900 & 39900 & 39900 & 39900 & 1 & 1 & 1 & 17 & 1 & 1 \\
& 35 & 43506 & 43506 & 43506 & 43506 & 43506 & 43506 & 3 & 1 & 1 & 31 & 1 & 1 \\
& 36 & 47323 & 47323 & 47323 & 47323 & 47323 & 47323 & 1 & 1 & 1 & 80 & 2 & 1 \\
& 37 & 51357 & 51357 & 51357 & 51357 & 51357 & 51357 & 2 & 2 & 2 & 17 & 5 & 1 \\
& 38 & 55614 & 55614 & 55614 & 55614 & 55614 & 55614 & 4 & 4 & 2 & 39 & 3 & 1 \\
& 39 & 60100 & 60100 & 60100 & 60100 & 60100 & 60100 & 3 & 3 & 3 & 26 & 26 & 1 \\
& 40 & 64821 & 64821 & 64821 & 64821 & 64821 & 64821 & 3 & 3 & 3 & 22 & 9 & 3 \\
& 41 & 69783 & 69783 & 69783 & 69783 & 69783 & 69783 & 1 & 1 & 1 & 59 & 59 & 1 \\
& 42 & 74992 & 74992 & 74992 & 74992 & 74992 & 74992 & 1 & 1 & 1 & 28 & 27 & 1 \\
& 43 & 80454 & 80454 & 80454 & 80454 & 80454 & 80454 & 2 & 2 & 2 & 121 & 13 & 2 \\
& 44 & 86175 & 86175 & 86175 & 86175 & 86175 & 86175 & 4 & 4 & 4 & 85 & 73 & 3 \\
& 45 & 92161 & 92161 & 92161 & 92161 & 92161 & 92161 & 6 & 6 & 6 & 157 & 67 & 2 \\
& 46 & 98418 & 98418 & 98418 & 98418 & 98418 & 98418 & 3 & 3 & 3 & 173 & 49 & 3 \\
& 47 & 104952 & 104952 & 104952 & 104952 & 104952 & 104952 & 3 & 3 & 3 & 121 & 51 & 17 \\
& 48 & 111769 & 111769 & 111769 & 111769 & 111769 & 111769 & 1 & 1 & 1 & 167 & 94 & 1 \\
& 49 & 118875 & 118875 & 118875 & 118875 & 118875 & 118875 & 3 & 3 & 3 & 84 & 84 & 30 \\
& 50 & 126276 & 126276 & 126276 & 126276 & 126276 & 126276 & 3 & 3 & 3 & 167 & 92 & 2 \\
    
    \midrule
    \hline \multicolumn{14}{|r|}{{Continued on next page}} \\ \hline
  \end{longtable}
\end{table}

\begin{table}[]
  \centering
  \footnotesize
  \begin{longtable}{llrrr|rrr|rrr|rrr}
    \toprule
 & & \multicolumn{3}{c}{RC2} & \multicolumn{3}{c}{FM} & \multicolumn{3}{c}{LSU} & \multicolumn{3}{c}{torchmSAT}\\
    \cmidrule(r){3-14}
    & & 1m & 5m & 10m & 1m & 5m & 10m & 1m  & 5m & 10m & 1m  & 5m & 10m \\
    \midrule
    \endhead
    PAR & 1 & 1 & 1 & 1 & 1 & 1 & 1 & 1 & 1 & 1 & 1 & 1 & 1 \\
& 2 & 1 & 1 & 1 & 1 & 1 & 1 & 1 & 1 & 1 & 1 & 1 & 1 \\
& 3 & 1 & 1 & 1 & 1 & 1 & 1 & 1 & 1 & 1 & 1 & 1 & 1 \\
& 4 & 1 & 1 & 1 & 1 & 1 & 1 & 1 & 1 & 1 & 3 & 1 & 1 \\
& 5 & 1 & 1 & 1 & 1 & 1 & 1 & 1 & 1 & 1 & 3 & 3 & 3 \\
& 6 & 1 & 1 & 1 & 1 & 1 & 1 & 1 & 1 & 1 & 5 & 3 & 3 \\
& 7 & 1 & 1 & 1 & 1380 & 1 & 1 & 1 & 1 & 1 & 7 & 7 & 7 \\
& 8 & 2057 & 2057 & 2057 & 2057 & 2057 & 2057 & 1 & 1 & 1 & 7 & 7 & 7 \\
& 9 & 2926 & 2926 & 2926 & 2926 & 2926 & 2926 & 1 & 1 & 1 & 5 & 5 & 5 \\
& 10 & 4011 & 4011 & 4011 & 4011 & 4011 & 4011 & 1 & 1 & 1 & 9 & 9 & 9 \\
& 11 & 5336 & 5336 & 5336 & 5336 & 5336 & 5336 & 1 & 1 & 1 & 11 & 11 & 9 \\
& 12 & 6925 & 6925 & 6925 & 6925 & 6925 & 6925 & 1 & 1 & 1 & 15 & 10 & 10 \\
& 13 & 8802 & 8802 & 8802 & 8802 & 8802 & 8802 & 1 & 1 & 1 & 17 & 11 & 11 \\
& 14 & 10991 & 10991 & 10991 & 10991 & 10991 & 10991 & 1 & 1 & 1 & 17 & 17 & 15 \\
& 15 & 13516 & 13516 & 13516 & 13516 & 13516 & 13516 & 1 & 1 & 1 & 19 & 17 & 17 \\
& 16 & 16401 & 16401 & 16401 & 16401 & 16401 & 16401 & 1 & 1 & 1 & 23 & 19 & 19 \\
& 17 & 19670 & 19670 & 19670 & 19670 & 19670 & 19670 & 1 & 1 & 1 & 23 & 23 & 23 \\
& 18 & 23347 & 23347 & 23347 & 23347 & 23347 & 23347 & 1 & 1 & 1 & 32 & 28 & 25 \\
& 19 & 27456 & 27456 & 27456 & 27456 & 27456 & 27456 & 1 & 1 & 1 & 38 & 32 & 28 \\
& 20 & 32021 & 32021 & 32021 & 32021 & 32021 & 32021 & 1 & 1 & 1 & 134 & 33 & 31 \\
& 21 & 37066 & 37066 & 37066 & 37066 & 37066 & 37066 & 1 & 1 & 1 & 4553 & 30 & 30 \\
& 22 & 42615 & 42615 & 42615 & 42615 & 42615 & 42615 & 1 & 1 & 1 & 5085 & 34 & 34 \\
& 23 & 48692 & 48692 & 48692 & 48692 & 48692 & 48692 & 1 & 1 & 1 & 5745 & 31 & 31 \\
& 24 & 55321 & 55321 & 55321 & 55321 & 55321 & 55321 & 1 & 1 & 1 & 6344 & 38 & 38 \\
& 25 & 62526 & 62526 & 62526 & 62526 & 62526 & 62526 & 1 & 1 & 1 & 7490 & 45 & 40 \\
& 26 & 70331 & 70331 & 70331 & 70331 & 70331 & 70331 & 1 & 1 & 1 & 8380 & 72 & 46 \\
& 27 & 78760 & 78760 & 78760 & 78760 & 78760 & 78760 & 1 & 1 & 1 & 9294 & 51 & 38 \\
& 28 & 87837 & 87837 & 87837 & 87837 & 87837 & 87837 & 1 & 1 & 1 & 10574 & 107 & 55 \\
& 29 & 97586 & 97586 & 97586 & 97586 & 97586 & 97586 & 1 & 1 & 1 & 11300 & 89 & 73 \\
& 30 & 108031 & 108031 & 108031 & 108031 & 108031 & 108031 & 1 & 1 & 1 & 12677 & 4705 & 53 \\
& 31 & 119196 & 119196 & 119196 & 119196 & 119196 & 119196 & 1 & 1 & 1 & 14268 & 4693 & 44 \\
& 32 & 131105 & 131105 & 131105 & 131105 & 131105 & 131105 & 1 & 1 & 1 & 15063 & 15063 & 126 \\
& 33 & 143782 & 143782 & 143782 & 143782 & 143782 & 143782 & 1 & 1 & 1 & 17086 & 17086 & 258 \\
& 34 & 157251 & 157251 & 157251 & 157251 & 157251 & 157251 & 1 & 1 & 1 & 18917 & 18917 & 466 \\
& 35 & 171536 & 171536 & 171536 & 171536 & 171536 & 171536 & 1 & 1 & 1 & 20864 & 20864 & 1249 \\
& 36 & 186661 & 186661 & 186661 & 186661 & 186661 & 186661 & 1 & 1 & 1 & 21759 & 21759 & 21759 \\
& 37 & 202650 & 202650 & 202650 & 202650 & 202650 & 202650 & 1 & 1 & 1 & 24458 & 24458 & 24458 \\
& 38 & 219527 & 219527 & 219527 & 219527 & 219527 & 219527 & 1 & 1 & 1 & 29072 & 26754 & 26754 \\
& 39 & 237316 & 237316 & 237316 & 237316 & 237316 & 237316 & 1 & 1 & 1 & 37080 & 28506 & 28506 \\
& 40 & 256041 & 256041 & 256041 & 256041 & 256041 & 256041 & 1 & 1 & 1 & 44488 & 30846 & 30846 \\
& 41 & 275726 & 275726 & 275726 & 275726 & 275726 & 275726 & 24 & 1 & 1 & 51984 & 32926 & 32926 \\
& 42 & 296395 & 296395 & 296395 & 296395 & 296395 & 296395 & 1 & 1 & 1 & 57826 & 34824 & 34824 \\
& 43 & 318072 & 318072 & 318072 & 318072 & 318072 & 318072 & 1 & 1 & 1 & 63468 & 37285 & 37285 \\
& 44 & 340781 & 340781 & 340781 & 340781 & 340781 & 340781 & 12 & 1 & 1 & 75878 & 41174 & 41174 \\
& 45 & 364546 & 364546 & 364546 & 364546 & 364546 & 364546 & 92 & 1 & 1 & 92245 & 43744 & 43744 \\
& 46 & 389391 & 389391 & 389391 & 389391 & 389391 & 389391 & 1 & 1 & 1 & 111230 & 46042 & 46042 \\
& 47 & 415340 & 415340 & 415340 & 415340 & 415340 & 415340 & 3 & 1 & 1 & 130513 & 49045 & 49045 \\
& 48 & 442417 & 442417 & 442417 & 442417 & 442417 & 442417 & 58 & 1 & 1 & 157500 & 52467 & 52467 \\
& 49 & 470646 & 470646 & 470646 & 470646 & 470646 & 470646 & 77 & 1 & 1 & 174228 & 55854 & 55854 \\
& 50 & 500051 & 500051 & 500051 & 500051 & 500051 & 500051 & 83 & 1 & 1 & 173900 & 60406 & 60406 \\
    
    \midrule
    \hline \multicolumn{14}{|r|}{{Continued on next page}} \\ \hline
  \end{longtable}
\end{table}

\begin{table}[]
  \centering
  \footnotesize
  \begin{longtable}{llrrr|rrr|rrr|rrr}
    \toprule
 & & \multicolumn{3}{c}{RC2} & \multicolumn{3}{c}{FM} & \multicolumn{3}{c}{LSU} & \multicolumn{3}{c}{torchmSAT}\\
    \cmidrule(r){3-14}
    & & 1m & 5m & 10m & 1m & 5m & 10m & 1m  & 5m & 10m & 1m  & 5m & 10m \\
    \midrule
    \endhead
    PHP & 1 & 1 & 1 & 1 & 1 & 1 & 1 & 1 & 1 & 1 & 1 & 1 & 1 \\
& 2 & 1 & 1 & 1 & 1 & 1 & 1 & 1 & 1 & 1 & 1 & 1 & 1 \\
& 3 & 1 & 1 & 1 & 1 & 1 & 1 & 1 & 1 & 1 & 1 & 1 & 1 \\
& 4 & 1 & 1 & 1 & 1 & 1 & 1 & 1 & 1 & 1 & 1 & 1 & 1 \\
& 5 & 1 & 1 & 1 & 1 & 1 & 1 & 1 & 1 & 1 & 1 & 1 & 1 \\
& 6 & 1 & 1 & 1 & 1 & 1 & 1 & 1 & 1 & 1 & 1 & 1 & 1 \\
& 7 & 1 & 1 & 1 & 1 & 1 & 1 & 1 & 1 & 1 & 2 & 2 & 2 \\
& 8 & 1 & 1 & 1 & 1 & 1 & 1 & 1 & 1 & 1 & 2 & 2 & 2 \\
& 9 & 1 & 1 & 1 & 1 & 1 & 1 & 1 & 1 & 1 & 3 & 3 & 3 \\
& 10 & 561 & 1 & 1 & 561 & 561 & 561 & 1 & 1 & 1 & 3 & 3 & 3 \\
& 11 & 738 & 738 & 738 & 738 & 738 & 738 & 1 & 1 & 1 & 5 & 5 & 5 \\
& 12 & 949 & 949 & 949 & 949 & 949 & 949 & 1 & 1 & 1 & 4 & 4 & 4 \\
& 13 & 1197 & 1197 & 1197 & 1197 & 1197 & 1197 & 1 & 1 & 1 & 7 & 6 & 6 \\
& 14 & 1485 & 1485 & 1485 & 1485 & 1485 & 1485 & 1 & 1 & 1 & 9 & 8 & 8 \\
& 15 & 1816 & 1816 & 1816 & 1816 & 1816 & 1816 & 1 & 1 & 1 & 10 & 8 & 8 \\
& 16 & 2193 & 2193 & 2193 & 2193 & 2193 & 2193 & 1 & 1 & 1 & 10 & 10 & 10 \\
& 17 & 2619 & 2619 & 2619 & 2619 & 2619 & 2619 & 1 & 1 & 1 & 13 & 10 & 10 \\
& 18 & 3097 & 3097 & 3097 & 3097 & 3097 & 3097 & 1 & 1 & 1 & 15 & 11 & 11 \\
& 19 & 3630 & 3630 & 3630 & 3630 & 3630 & 3630 & 1 & 1 & 1 & 16 & 12 & 12 \\
& 20 & 4221 & 4221 & 4221 & 4221 & 4221 & 4221 & 1 & 1 & 1 & 13 & 13 & 13 \\
& 21 & 4873 & 4873 & 4873 & 4873 & 4873 & 4873 & 1 & 1 & 1 & 16 & 16 & 15 \\
& 22 & 5589 & 5589 & 5589 & 5589 & 5589 & 5589 & 1 & 1 & 1 & 20 & 19 & 19 \\
& 23 & 6372 & 6372 & 6372 & 6372 & 6372 & 6372 & 1 & 1 & 1 & 32 & 23 & 17 \\
& 24 & 7225 & 7225 & 7225 & 7225 & 7225 & 7225 & 1 & 1 & 1 & 23 & 23 & 22 \\
& 25 & 8151 & 8151 & 8151 & 8151 & 8151 & 8151 & 1 & 1 & 1 & 36 & 27 & 27 \\
& 26 & 9153 & 9153 & 9153 & 9153 & 9153 & 9153 & 1 & 1 & 1 & 37 & 37 & 30 \\
& 27 & 10234 & 10234 & 10234 & 10234 & 10234 & 10234 & 1 & 1 & 1 & 34 & 34 & 34 \\
& 28 & 11397 & 11397 & 11397 & 11397 & 11397 & 11397 & 1 & 1 & 1 & 45 & 31 & 31 \\
& 29 & 12645 & 12645 & 12645 & 12645 & 12645 & 12645 & 1 & 1 & 1 & 43 & 43 & 35 \\
& 30 & 13981 & 13981 & 13981 & 13981 & 13981 & 13981 & 1 & 1 & 1 & 54 & 54 & 43 \\
& 31 & 15408 & 15408 & 15408 & 15408 & 15408 & 15408 & 1 & 1 & 1 & 106 & 47 & 42 \\
& 32 & 16929 & 16929 & 16929 & 16929 & 16929 & 16929 & 1 & 1 & 1 & 75 & 55 & 55 \\
& 33 & 18547 & 18547 & 18547 & 18547 & 18547 & 18547 & 1 & 1 & 1 & 64 & 64 & 57 \\
& 34 & 20265 & 20265 & 20265 & 20265 & 20265 & 20265 & 1 & 1 & 1 & 130 & 49 & 49 \\
& 35 & 22086 & 22086 & 22086 & 22086 & 22086 & 22086 & 1 & 1 & 1 & 84 & 80 & 66 \\
& 36 & 24013 & 24013 & 24013 & 24013 & 24013 & 24013 & 1 & 1 & 1 & 191 & 86 & 85 \\
& 37 & 26049 & 26049 & 26049 & 26049 & 26049 & 26049 & 1 & 1 & 1 & 538 & 117 & 103 \\
& 38 & 28197 & 28197 & 28197 & 28197 & 28197 & 28197 & 1 & 1 & 1 & 1651 & 70 & 70 \\
& 39 & 30460 & 30460 & 30460 & 30460 & 30460 & 30460 & 1 & 1 & 1 & 3596 & 111 & 100 \\
& 40 & 32841 & 32841 & 32841 & 32841 & 32841 & 32841 & 1 & 1 & 1 & 3985 & 115 & 111 \\
& 41 & 35343 & 35343 & 35343 & 35343 & 35343 & 35343 & 1 & 1 & 1 & 4231 & 146 & 128 \\
& 42 & 37969 & 37969 & 37969 & 37969 & 37969 & 37969 & 1 & 1 & 1 & 4607 & 86 & 86 \\
& 43 & 40722 & 40722 & 40722 & 40722 & 40722 & 40722 & 1 & 1 & 1 & 4936 & 145 & 145 \\
& 44 & 43605 & 43605 & 43605 & 43605 & 43605 & 43605 & 1 & 1 & 1 & 5171 & 108 & 98 \\
& 45 & 46621 & 46621 & 46621 & 46621 & 46621 & 46621 & 1 & 1 & 1 & 5399 & 128 & 128 \\
& 46 & 49773 & 49773 & 49773 & 49773 & 49773 & 49773 & 1 & 1 & 1 & 5924 & 161 & 115 \\
& 47 & 53064 & 53064 & 53064 & 53064 & 53064 & 53064 & 1 & 1 & 1 & 6370 & 56 & 56 \\
& 48 & 56497 & 56497 & 56497 & 56497 & 56497 & 56497 & 1 & 1 & 1 & 7023 & 145 & 88 \\
& 49 & 60075 & 60075 & 60075 & 60075 & 60075 & 60075 & 1 & 1 & 1 & 7133 & 355 & 194 \\
& 50 & 63801 & 63801 & 63801 & 63801 & 63801 & 63801 & 1 & 1 & 1 & 7552 & 233 & 233 \\
    
    \midrule
  \end{longtable}
\end{table}

\newpage

\section{GPU Acceleration}
\label{appendix:gpu}
\begin{figure}[h]
\renewcommand{\tabcolsep}{0pt}
\centering
\begin{tabular}{cc}
\includegraphics[clip, scale=0.25, valign=m]{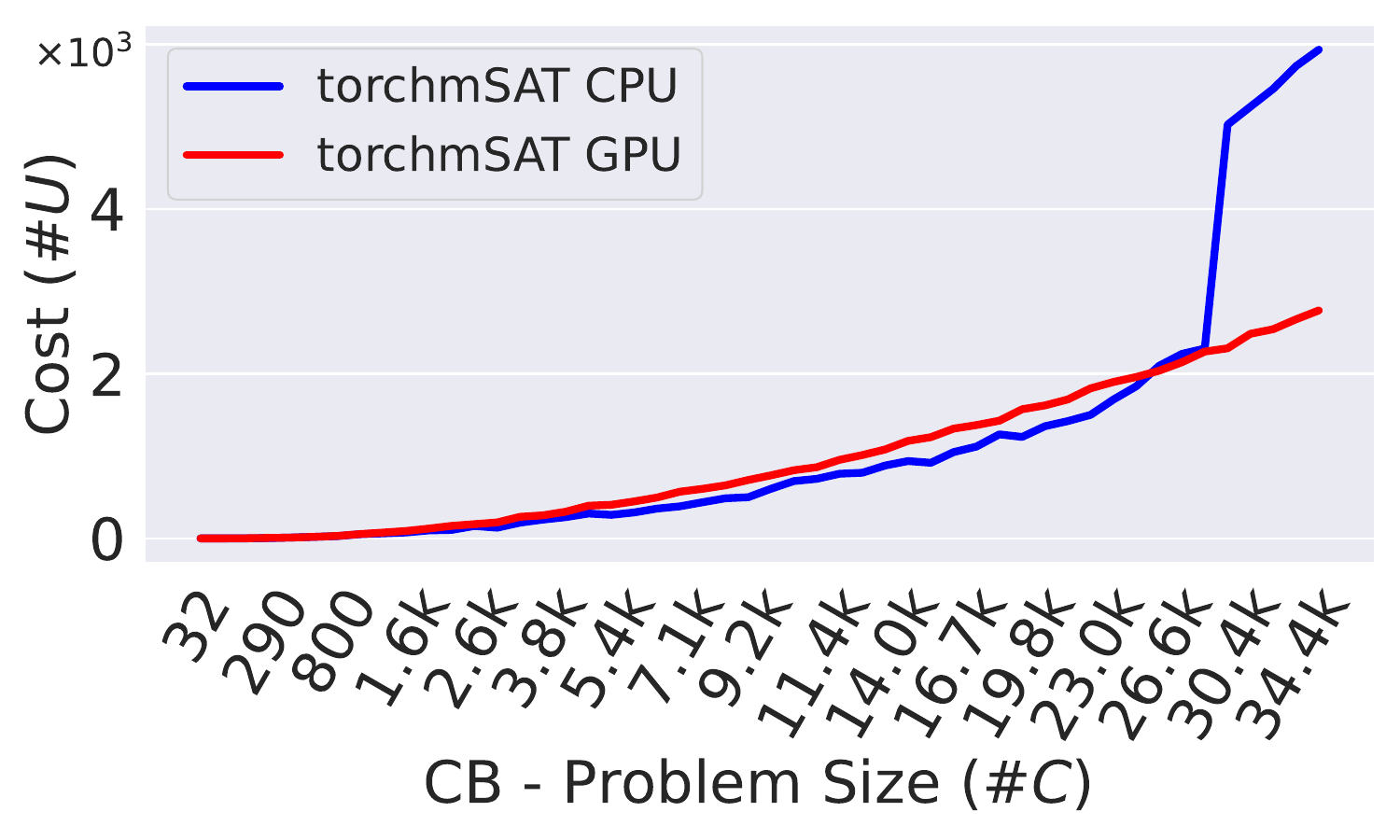} &
\includegraphics[clip, scale=0.25, valign=m]{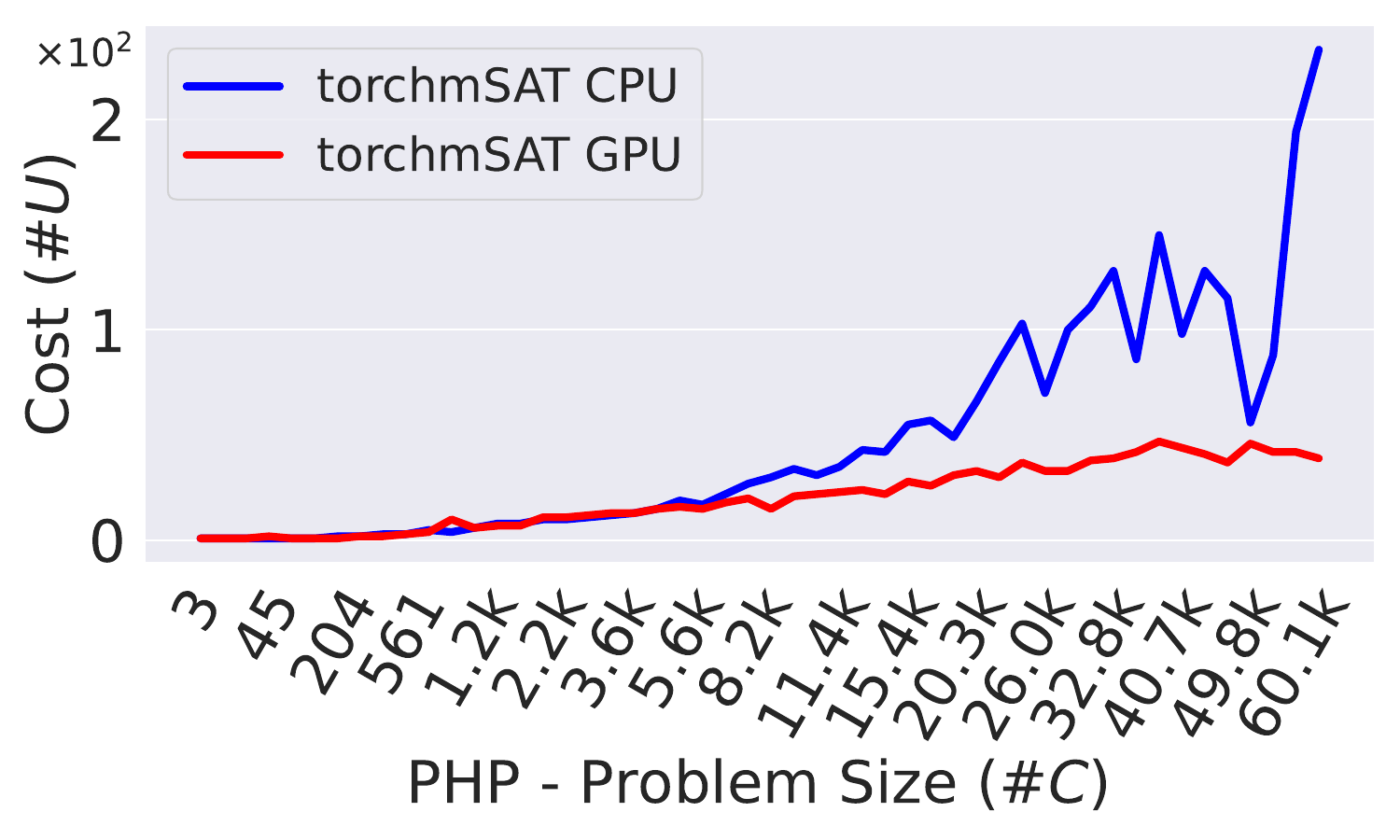} \\
\includegraphics[clip, scale=0.25, valign=m]{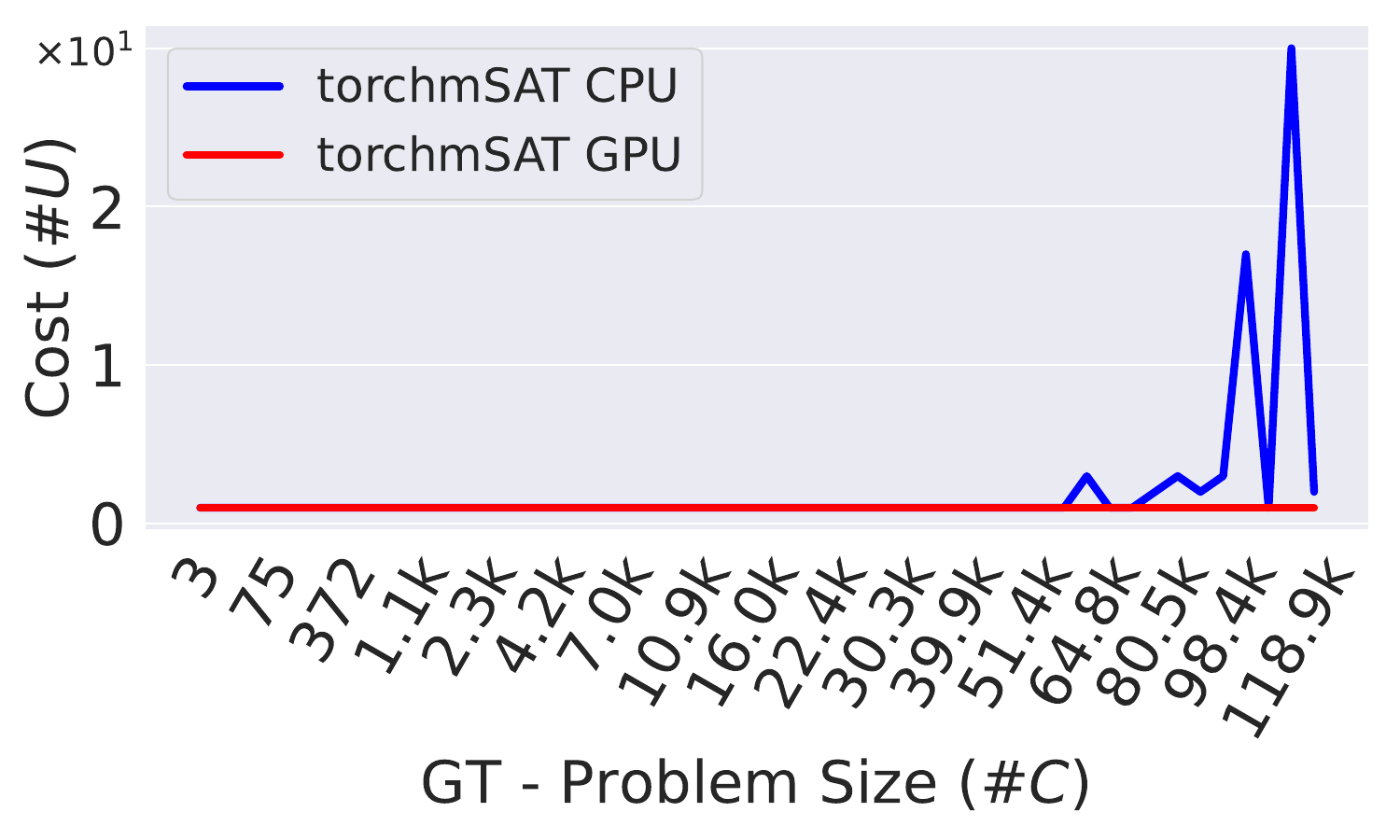} &
\includegraphics[clip, scale=0.25, valign=m]{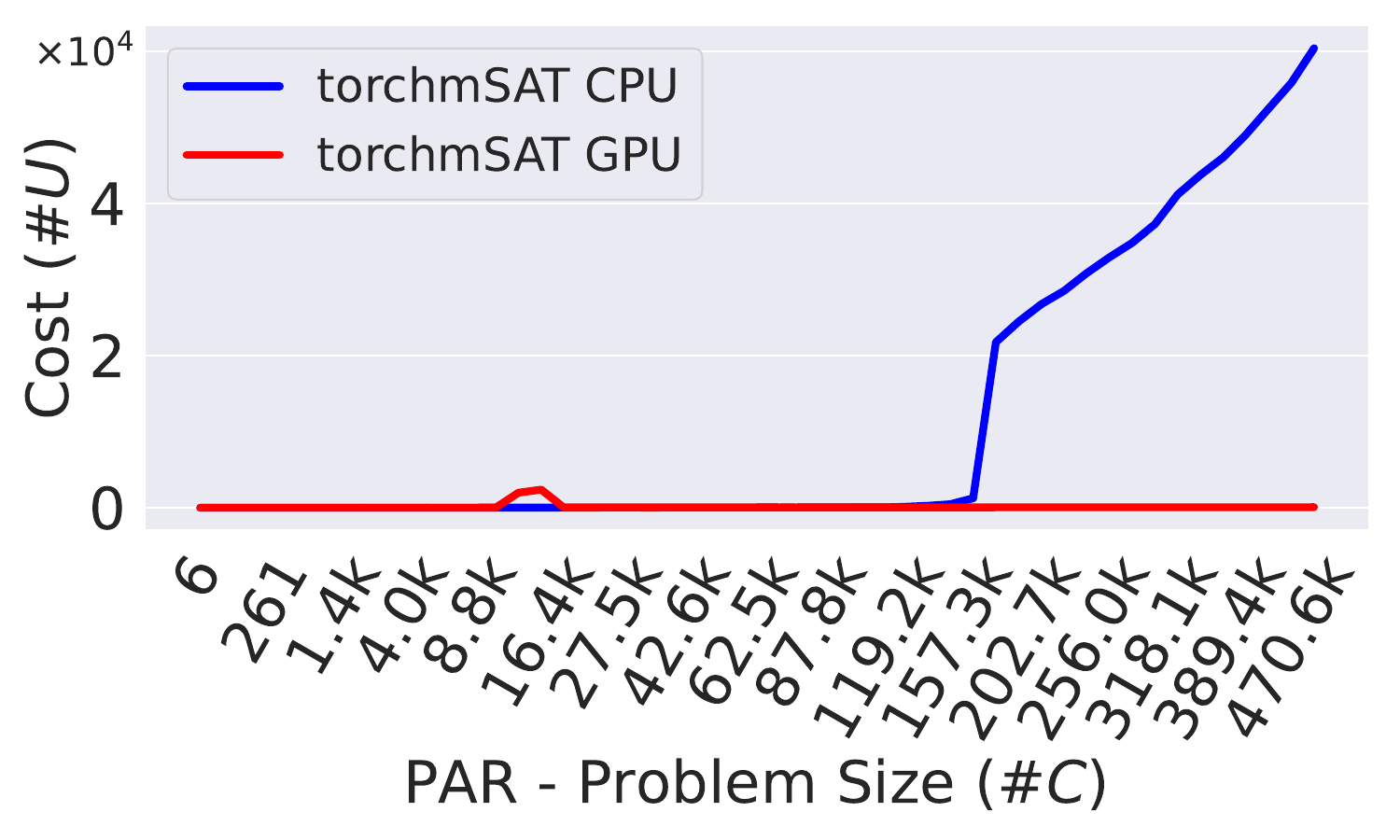}%
\end{tabular}
\caption{Running torchmSAT on CPU vs. GPU, where it is capable of taking advantage of GPU acceleration, and finds better MaxSAT solutions within the same time limit \textbf{(10mins)}.}
\label{fig:gpu-results-600}
\end{figure}

\begin{figure}[h]
\renewcommand{\tabcolsep}{0pt}
\centering
\begin{tabular}{cc}
\includegraphics[clip, scale=0.25, valign=m]{figs/neurips_cb_300_gpu.pdf} &
\includegraphics[clip, scale=0.25, valign=m]{figs/neurips_php_300_gpu.pdf} \\
\includegraphics[clip, scale=0.25, valign=m]{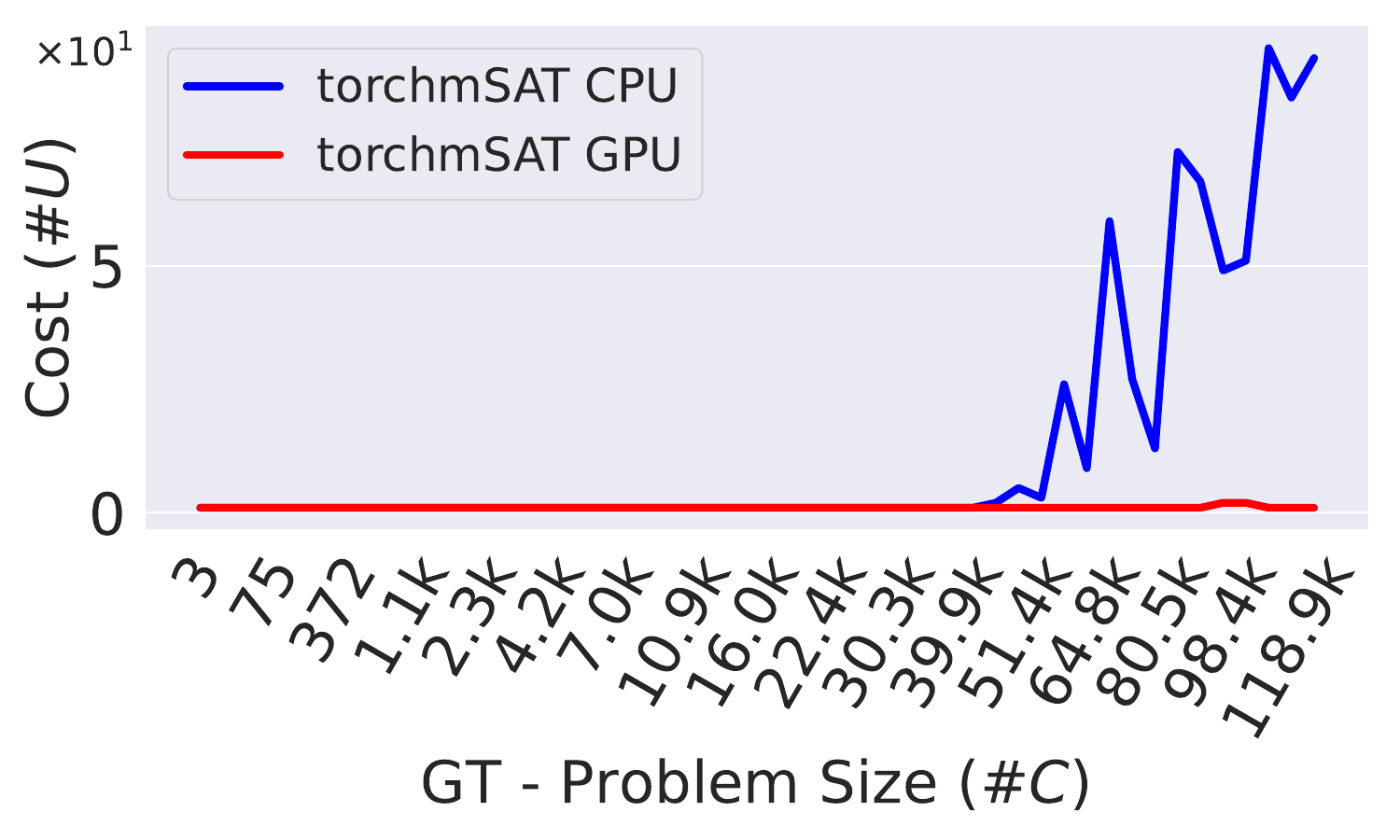} &
\includegraphics[clip, scale=0.25, valign=m]{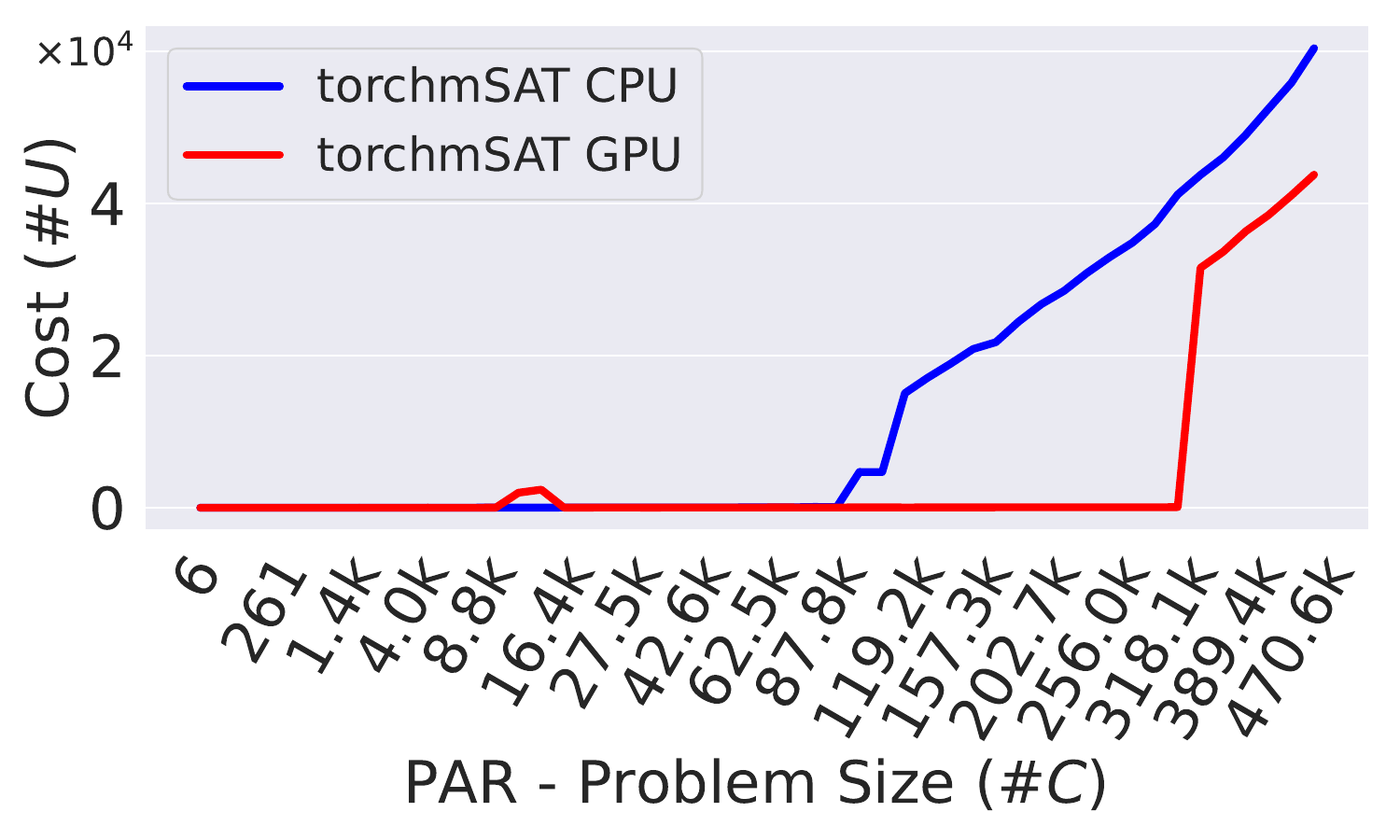}%
\end{tabular}
\caption{Running torchmSAT on CPU vs. GPU, where it is capable of taking advantage of GPU acceleration, and finds better MaxSAT solutions within the same time limit \textbf{(5mins)}.}
\label{fig:gpu-results-300}
\end{figure}

\begin{figure}[h]
\renewcommand{\tabcolsep}{0pt}
\centering
\begin{tabular}{cc}
\includegraphics[clip, scale=0.25, valign=m]{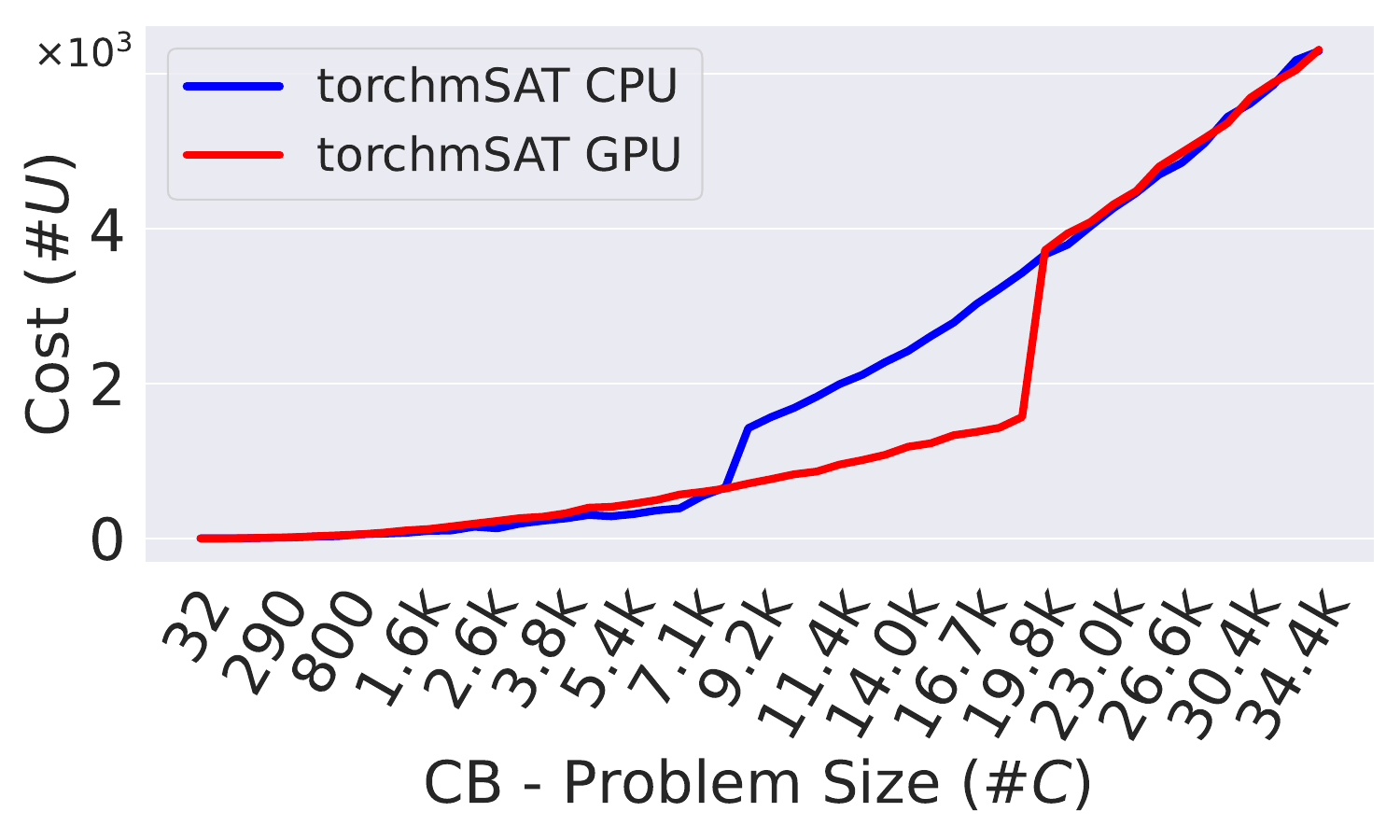} &
\includegraphics[clip, scale=0.25, valign=m]{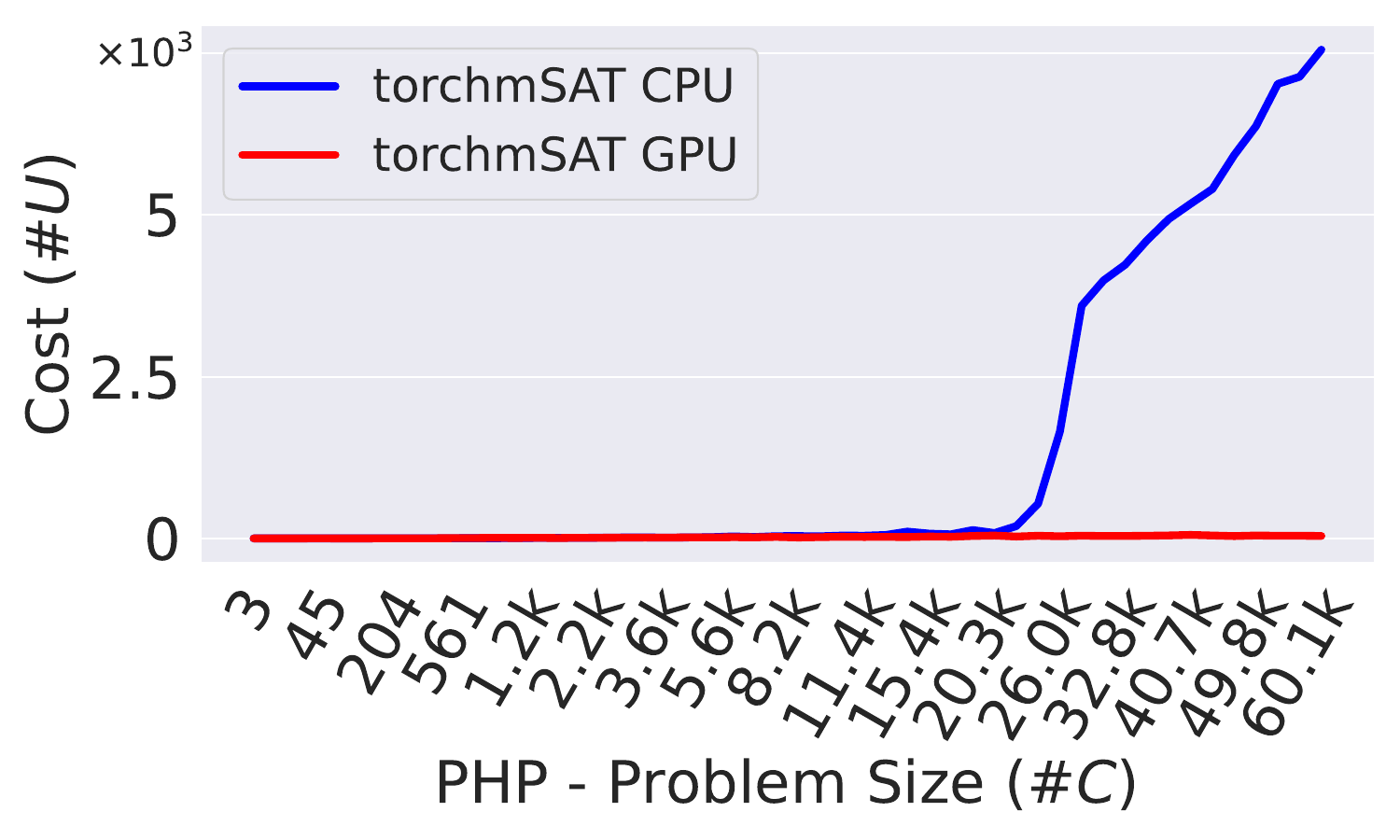} \\
\includegraphics[clip, scale=0.25, valign=m]{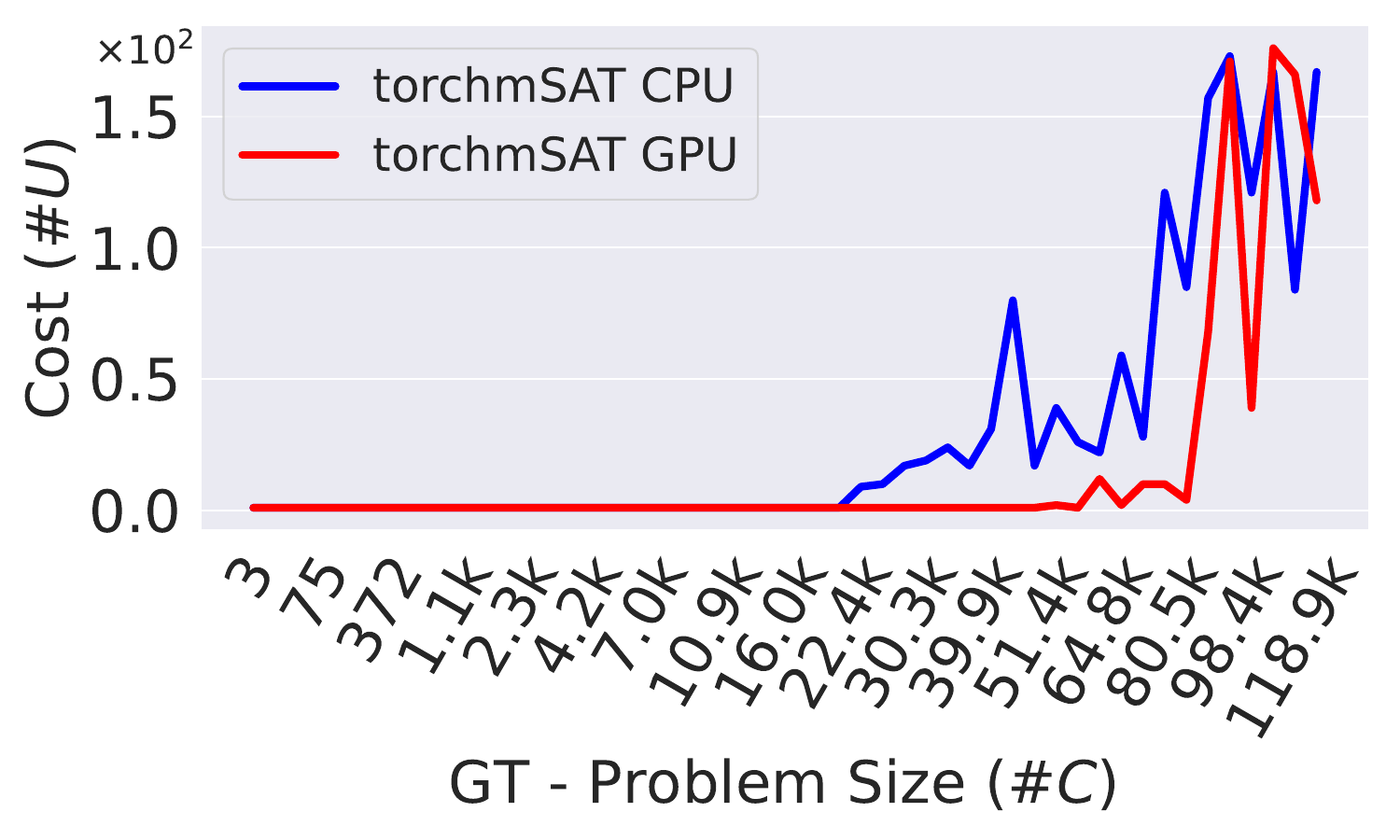} &
\includegraphics[clip, scale=0.25, valign=m]{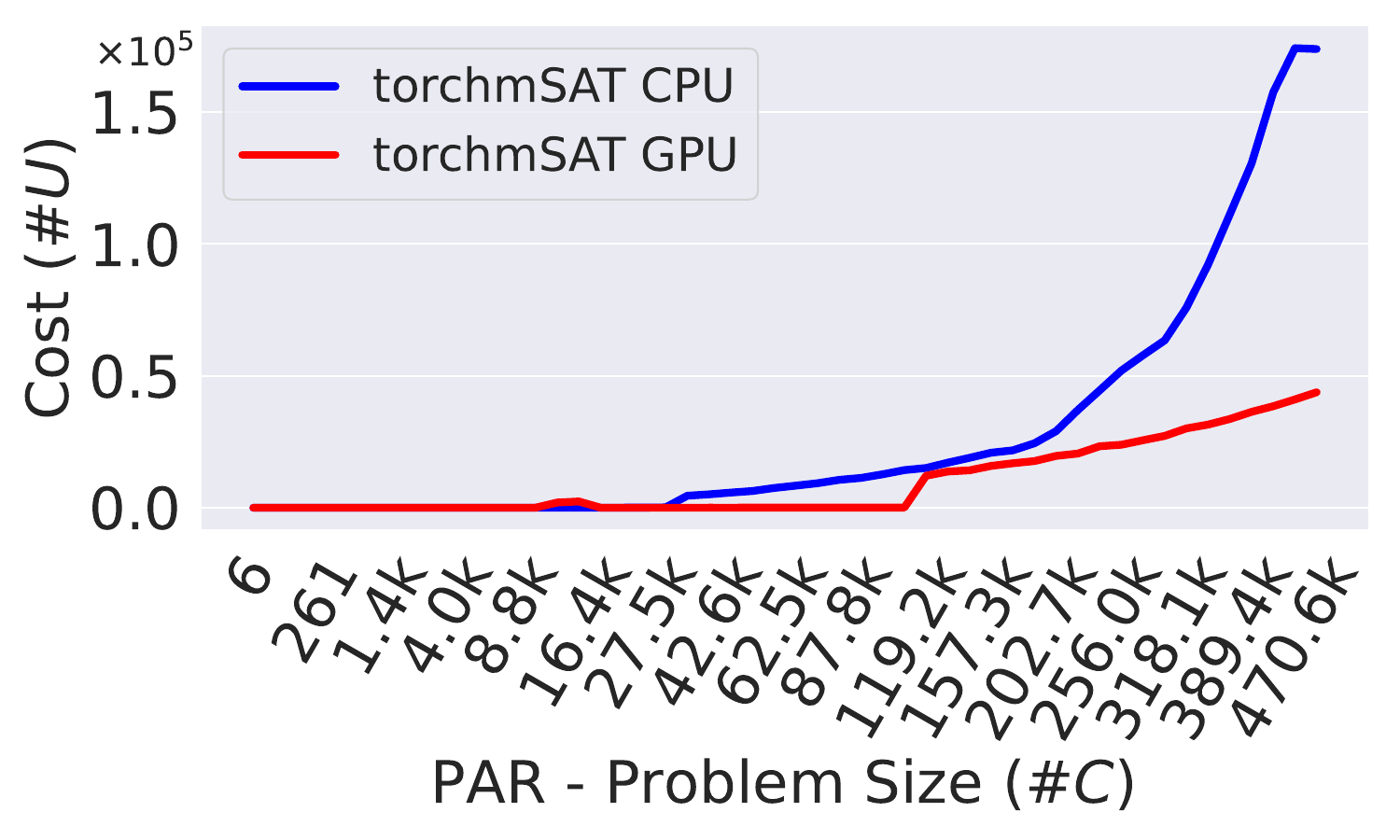}%
\end{tabular}
\caption{Running torchmSAT on CPU vs. GPU, where it is capable of taking advantage of GPU acceleration, and finds better MaxSAT solutions within the same time limit \textbf{(1min)}.}
\label{fig:gpu-results-60}
\end{figure}

The application of GPU acceleration in computational tasks offers remarkable advantages, particularly when it comes to large-scale computations, such as those involved in combinatorial optimization.
The parallel processing capabilities of GPUs allow for significant speed enhancements, enabling faster computation and consequently more rapid exploration of solution spaces.
For our torchmSAT method, which is reliant on neural networks and iterative computation, GPU acceleration can significantly enhance the efficiency of the forward-loss-backward loop, enabling more rapid exploration of the feasible region of variable assignments.
This implies that given a time limit, we can expect torchmSAT running on a GPU to find solutions of lower cost than it would when executed on a CPU.
Future exploration of GPU acceleration could bring about further advancements in MaxSAT solvers and other similar combinatorial optimization problems, leading to more efficient and effective solutions.
Figures \ref{fig:gpu-results-600}, \ref{fig:gpu-results-300} and \ref{fig:gpu-results-60} show the scalability of our method under different time limits.
As expected, the use of GPUs pays off on larger problem instances, especially when the time limit is tight.

\end{document}